%% file: arxiv.tex
\documentclass[conference, letterpaper]{IEEEtran}
\IEEEoverridecommandlockouts

\usepackage{cite}
\usepackage{amsmath,amssymb,amsfonts}
\usepackage[ruled,vlined]{algorithm2e}
\usepackage{algpseudocode}
\usepackage{graphicx}
\usepackage{textcomp}
\usepackage{xcolor}
\usepackage{caption}
\usepackage{subcaption}
\usepackage{array}
\usepackage{makecell}
\usepackage{dsfont}
\usepackage{threeparttable}
\usepackage{multirow}
\usepackage{wrapfig}
\usepackage[subtle]{savetrees}
\usepackage[hyphens]{url}
\usepackage{hyperref}
\hypersetup{breaklinks=true}
\def\BibTeX{{\rm B\kern-.05em{\sc i\kern-.025em b}\kern-.08em
    T\kern-.1667em\lower.7ex\hbox{E}\kern-.125emX}}
\usepackage{minted}
\usepackage{listings}
\lstdefinestyle{pythonstyle}{
    language=Python,
    basicstyle=\ttfamily\small,
    keywordstyle=\color{blue},
    commentstyle=\color{green!60!black},
    stringstyle=\color{red},
    numberstyle=\tiny\color{gray},
    numbers=left,
    numbersep=5pt,
    frame=single,
    breaklines=true,
    breakatwhitespace=true,
    tabsize=4,
    showstringspaces=false
}
\lstdefinestyle{promptstyle}{
    basicstyle=\ttfamily\small,
    frame=single,
    backgroundcolor=\color{gray!10},
    breaklines=true,
    columns=flexible
}
\usepackage[T1]{fontenc}  
\definecolor{delim}{RGB}{20,105,176}
\definecolor{numb}{RGB}{106, 109, 32}
\definecolor{string}{rgb}{0.64,0.08,0.08}

\lstdefinelanguage{json}{
    numbers=left,
    numberstyle=\small,
    frame=single,
    rulecolor=\color{black},
    showspaces=false,
    showtabs=false,
    breaklines=true,
    postbreak=\raisebox{0ex}[0ex][0ex]{\ensuremath{\color{gray}\hookrightarrow\space}},
    breakatwhitespace=true,
    basicstyle=\ttfamily\small,
    upquote=true,
    morestring=[b]",
    stringstyle=\color{string},
    literate=
     *{0}{{{\color{numb}0}}}{1}
      {1}{{{\color{numb}1}}}{1}
      {2}{{{\color{numb}2}}}{1}
      {3}{{{\color{numb}3}}}{1}
      {4}{{{\color{numb}4}}}{1}
      {5}{{{\color{numb}5}}}{1}
      {6}{{{\color{numb}6}}}{1}
      {7}{{{\color{numb}7}}}{1}
      {8}{{{\color{numb}8}}}{1}
      {9}{{{\color{numb}9}}}{1}
      {\{}{{{\color{delim}{\{}}}}{1}
      {\}}{{{\color{delim}{\}}}}}{1}
      {[}{{{\color{delim}{[}}}}{1}
      {]}{{{\color{delim}{]}}}}{1},
}

\makeatletter
\def\@begintheorem#1#2[#3]{%
  \trivlist
  \item[\hskip\labelsep\bfseries #1\ #2\ (#3).]%
  \setlength{\itemindent}{0em}
  \vspace{-0.5\topsep}}
\def\@opargbegintheorem#1#2#3{%
  \trivlist
  \item[\hskip\labelsep\bfseries #1\ #2\ (#3).]%
  \setlength{\itemindent}{0em}
  \vspace{-0.5\topsep}}
\makeatletter
\def\@endtheorem{\endtrivlist\vspace{-\topsep}}
\makeatother

\newenvironment{compactItemize}{\begin{list}{$\bullet$}
{\setlength{\topsep}{1mm}\setlength{\itemsep}{0.25mm}
\setlength{\parsep}{0.1mm}
\setlength{\itemindent}{0mm}\setlength{\partopsep}{0mm}
\setlength{\labelwidth}{15mm}
\setlength{\leftmargin}{4mm}}}{\end{list}}

\newtheorem{definition}{Definition}

\newtheorem{example}{Example}

\begin{document}
\bstctlcite{IEEEexample:BSTcontrol}
\pagestyle{plain}

\title{ScenicRules: An Autonomous Driving Benchmark with Multi-Objective Specifications and Abstract Scenarios \vspace*{-0.6cm}}

\author{
\IEEEauthorblockN{\vspace*{-0.6cm}}
\parbox{\textwidth}{\centering
Kevin Kai-Chun Chang\textsuperscript{1}, Ekin Beyazit\textsuperscript{2}, Alberto Sangiovanni-Vincentelli\textsuperscript{1}, Tichakorn Wongpiromsarn\textsuperscript{2}, and Sanjit A. Seshia\textsuperscript{1} \\
\textsuperscript{1}University of California, Berkeley \quad \textsuperscript{2}Iowa State University \\
\{kaichunchang, alberto, sseshia\}@berkeley.edu, \{ekin, nok\}@iastate.edu
}\vspace*{-1.2cm}
\thanks{Kevin Kai-Chun Chang and Ekin Beyazit contributed equally. This work was supported in part by NSF grants CNS-2211141, CNS-2141153, and 2303564, by the DARPA TIAMAT program, by the California Partners for Advanced Transportation Technology (PATH) Future Mobility Research Center (FMRC), and by Nissan under the iCyPhy center. The Berkeley authors gratefully acknowledge feedback from Prof. Scott Moura and Dr. Alex Kurzhanskiy.}
}

\maketitle

\thispagestyle{plain}

\begin{abstract}
Developing autonomous driving systems for complex traffic environments requires balancing multiple objectives, such as avoiding collisions, obeying traffic rules, and making efficient progress. In many situations, these objectives cannot be satisfied simultaneously, and explicit priority relations naturally arise. Also, driving rules require context, so it is important to formally model the environment scenarios within which such rules apply. Existing benchmarks for evaluating autonomous vehicles lack such combinations of multi-objective prioritized rules and formal environment models. In this work, we introduce \textit{ScenicRules}, a benchmark for evaluating autonomous driving systems in stochastic environments under prioritized multi-objective specifications. We first formalize a diverse set of objectives to serve as quantitative evaluation metrics. Next, we design a Hierarchical Rulebook framework that encodes multiple objectives and their priority relations in an interpretable and adaptable manner. We then construct a compact yet representative collection of scenarios spanning diverse driving contexts and near-accident situations, formally modeled in the Scenic language. Experimental results show that our formalized objectives and Hierarchical Rulebooks align well with human driving judgments and that our benchmark effectively exposes agent failures with respect to the prioritized objectives. Our benchmark can be accessed at \url{https://github.com/BerkeleyLearnVerify/ScenicRules/}.
\end{abstract}


\input{sections_arxiv/01_introduction}
\input{sections_arxiv/02_related_work}
\input{sections_arxiv/03_preliminaries}
\input{sections_arxiv/04_benchmark}
\input{sections_arxiv/05_experiments}
\input{sections_arxiv/06_conclusion}


{\footnotesize{
\bibliographystyle{IEEEtran}
\bibliography{ref}
}}

\newpage
\input{sections_arxiv/appendix}

\end{document}

%% file: sections_arxiv/01_introduction.tex

\section{Introduction}\label{sec:intro}

\begin{figure*}[tb]
    \centering
    \begin{subfigure}[t]{\textwidth}
        \centering
        \includegraphics[width=0.19\textwidth]{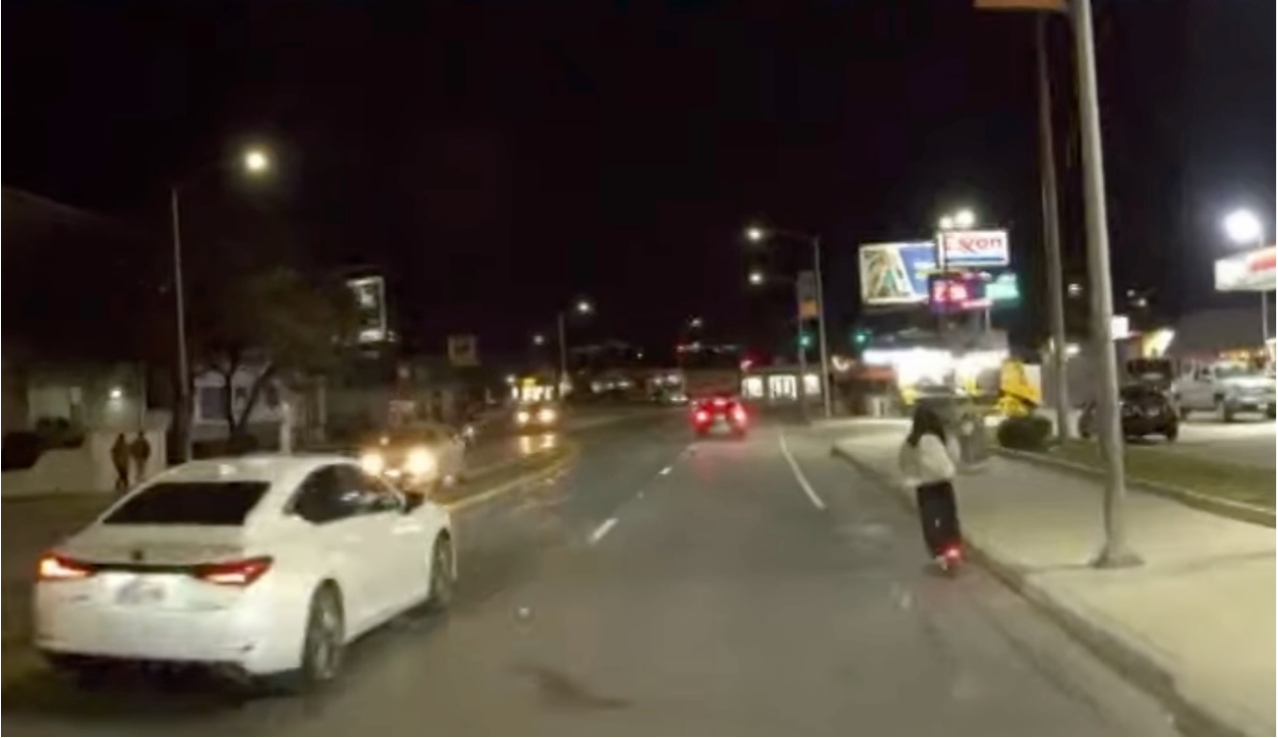}
        \includegraphics[width=0.19\textwidth]{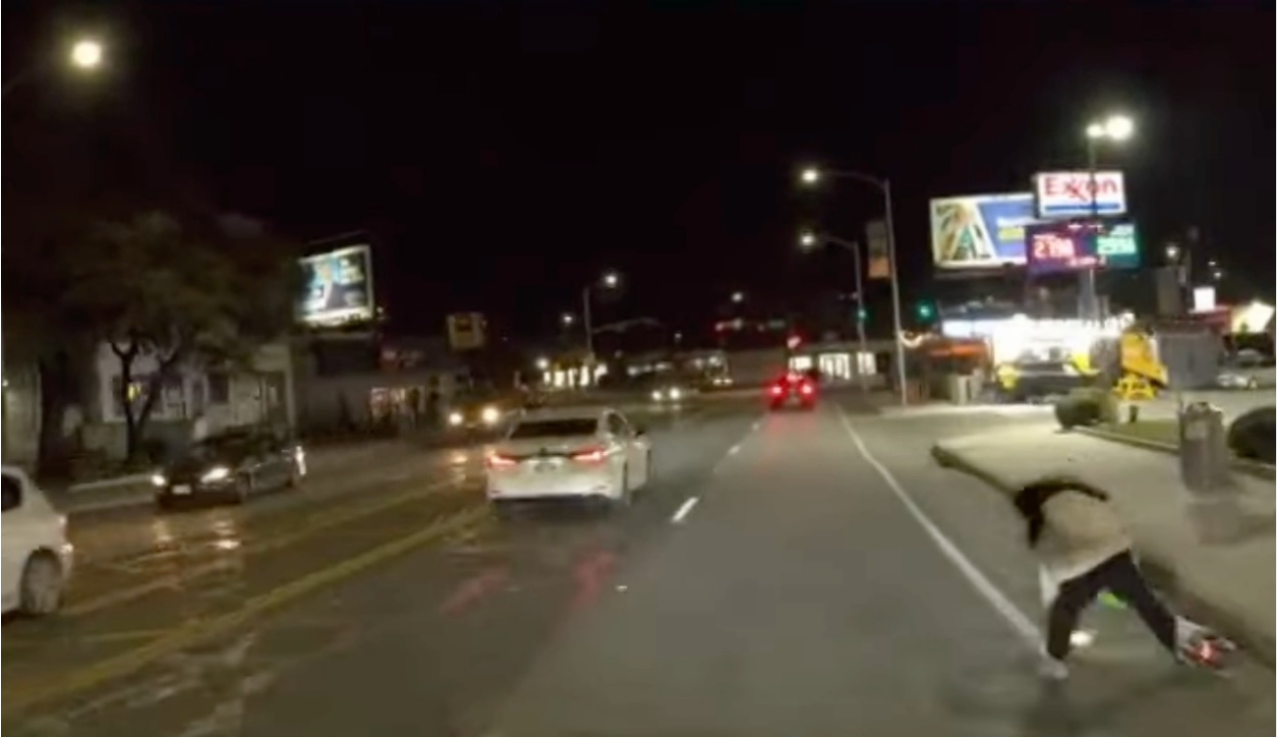}
        \includegraphics[width=0.19\textwidth]{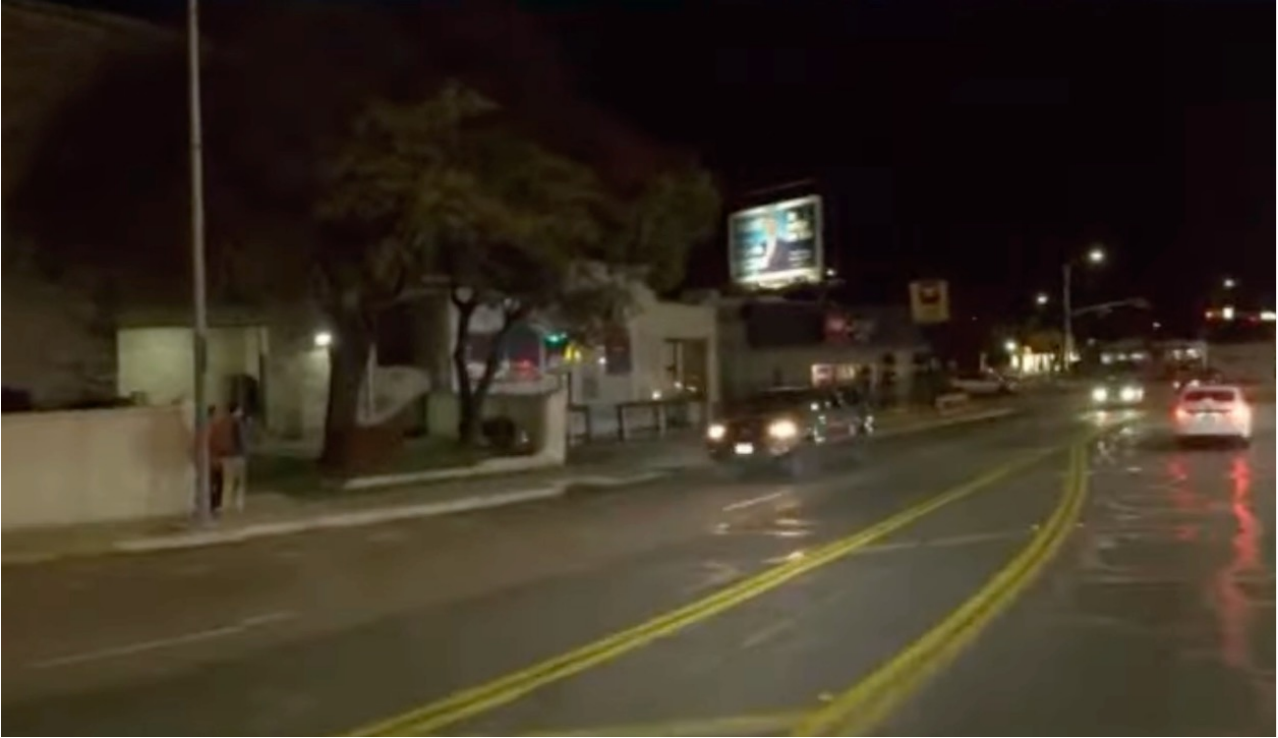}
        \includegraphics[width=0.19\textwidth]{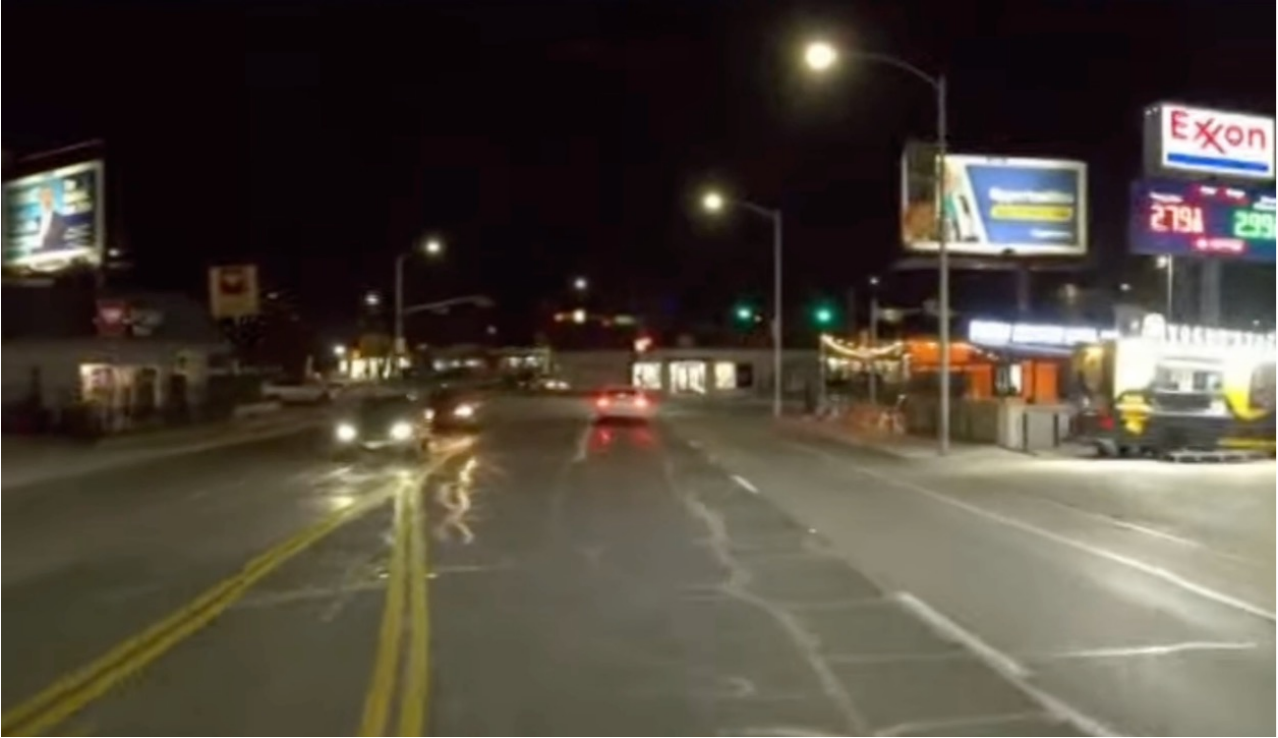}
        \includegraphics[width=0.19\textwidth]{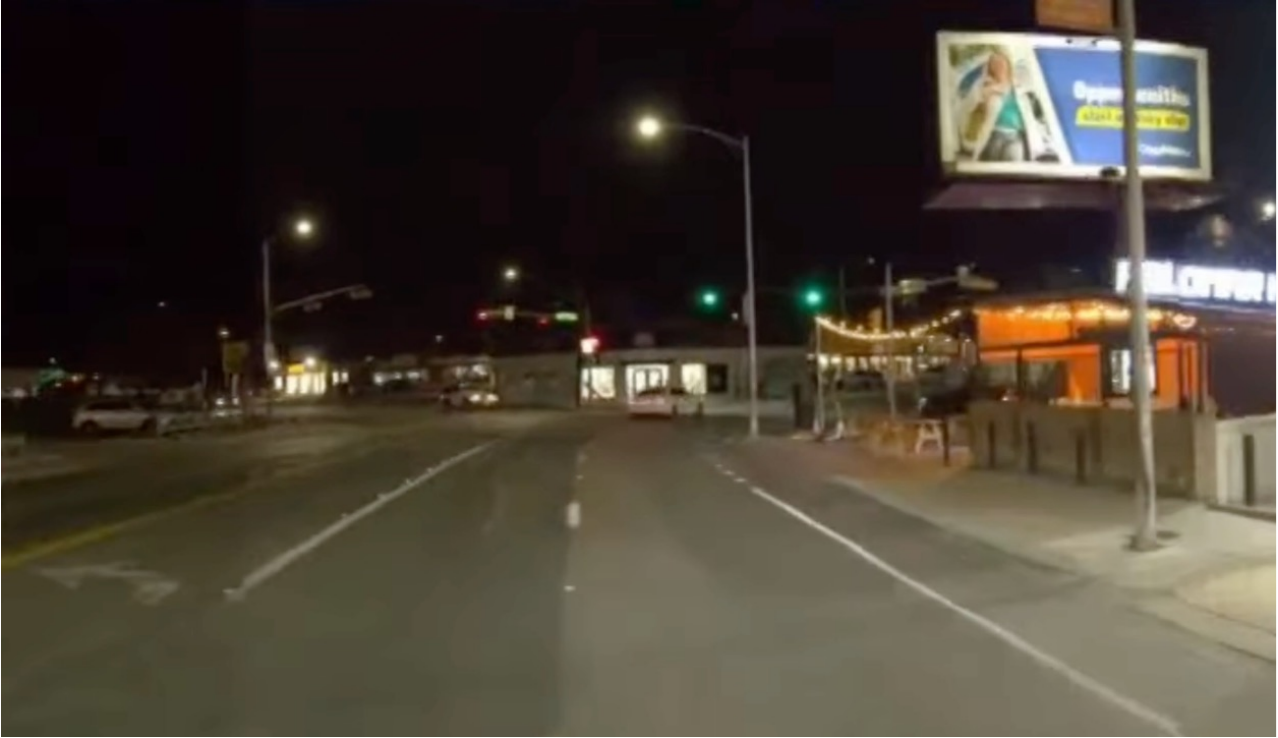}
        \caption{A Waymo autonomous vehicle prioritizes \textit{collision avoidance} over \textit{lane keeping}~\cite{kxan2024waymo}.}
        \label{fig:waymo_example}
        \vspace*{2mm}
    \end{subfigure}
    \begin{subfigure}[t]{\textwidth}
        \centering
        \includegraphics[width=0.19\textwidth]{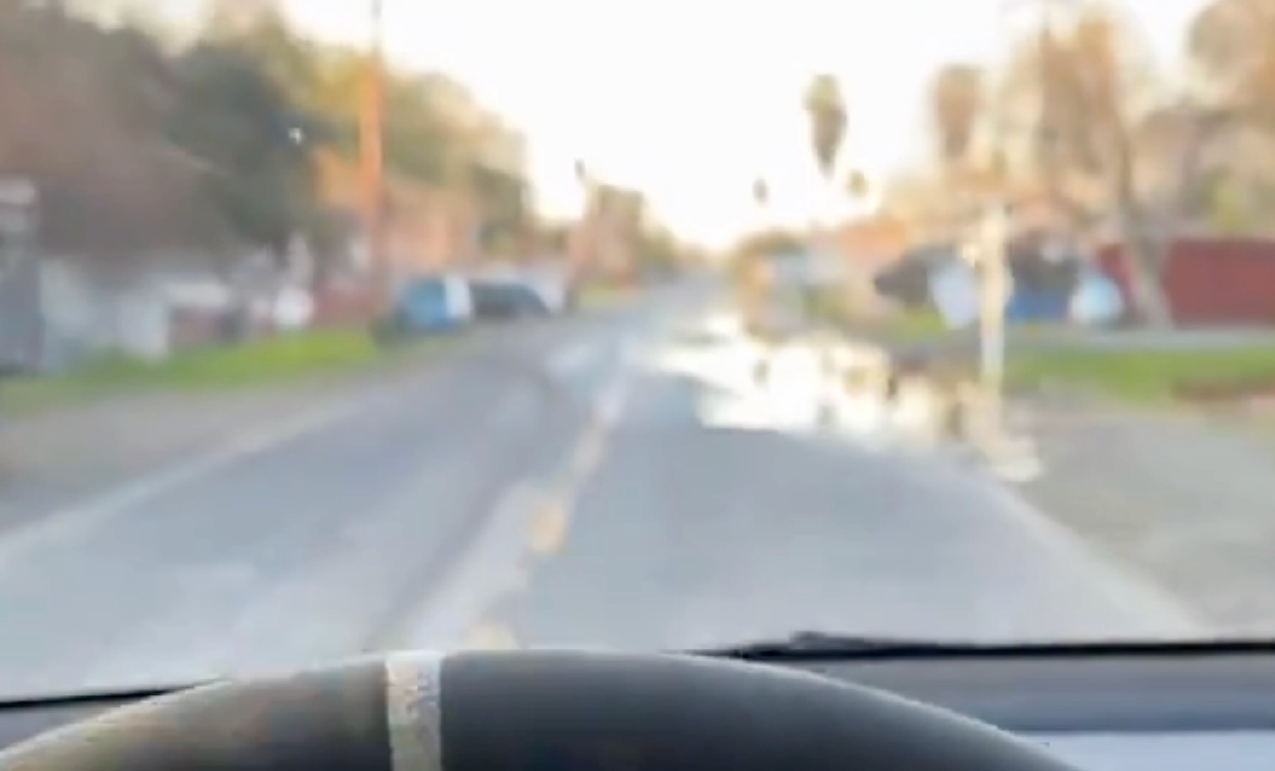}
        \includegraphics[width=0.19\textwidth]{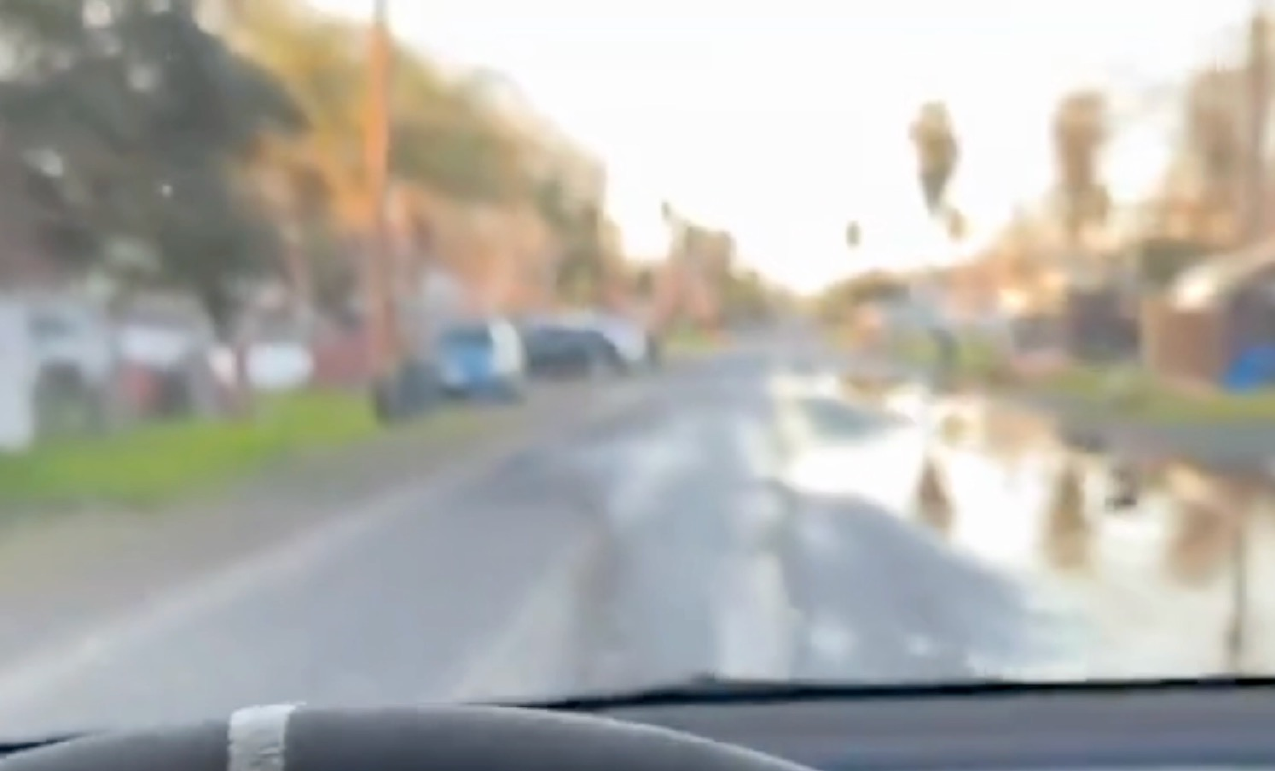}
        \includegraphics[width=0.19\textwidth]{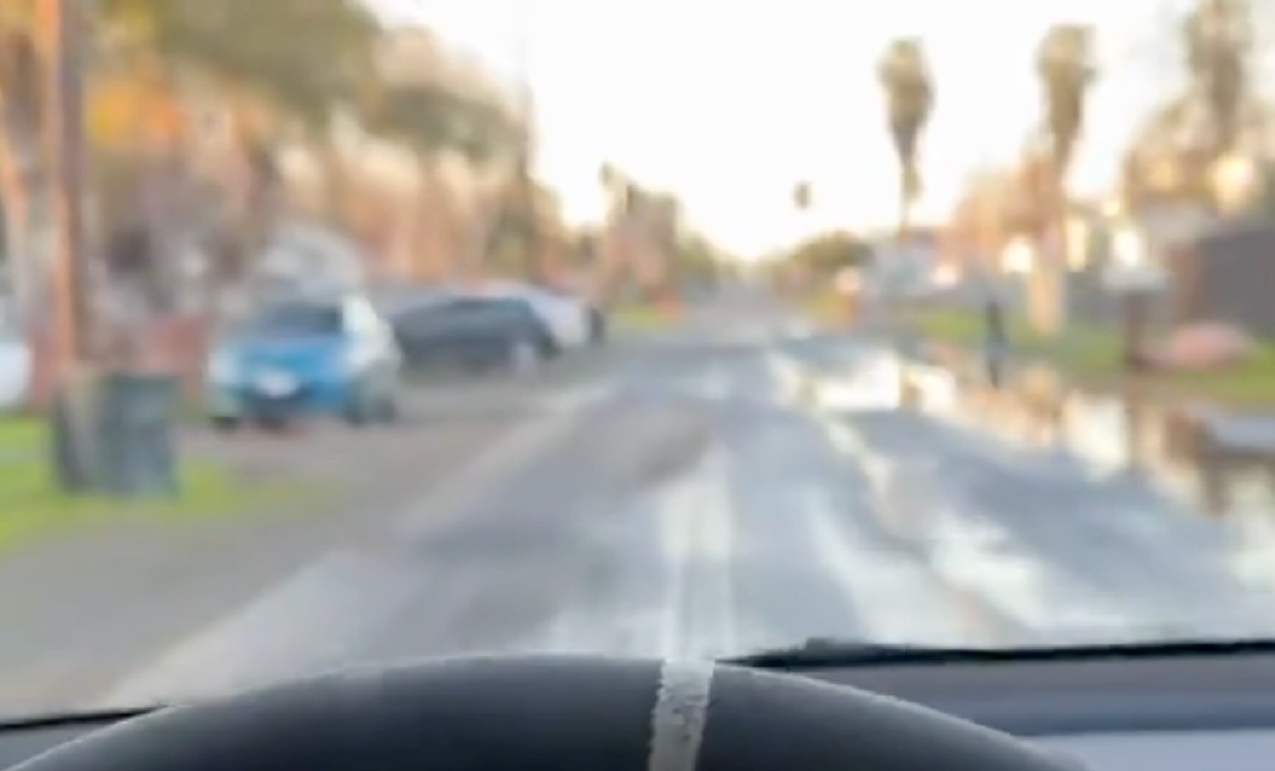}
        \includegraphics[width=0.19\textwidth]{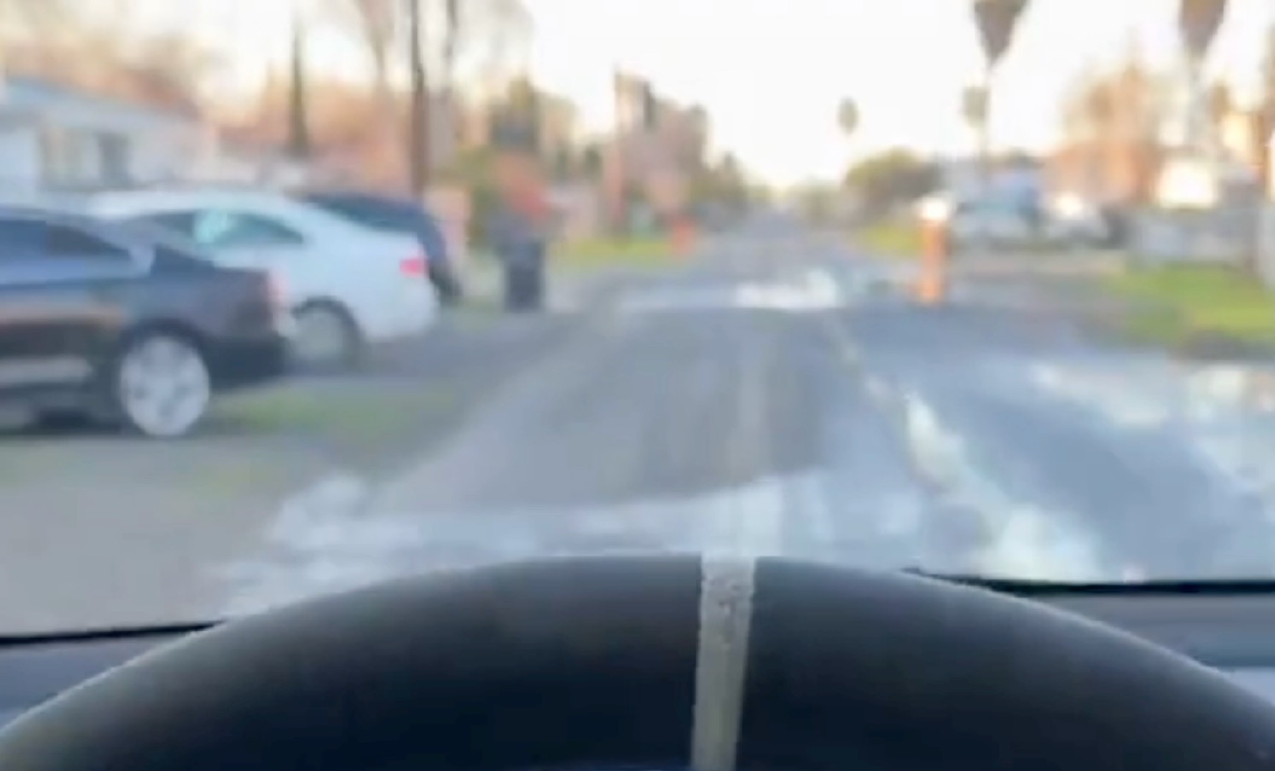}
        \includegraphics[width=0.19\textwidth]{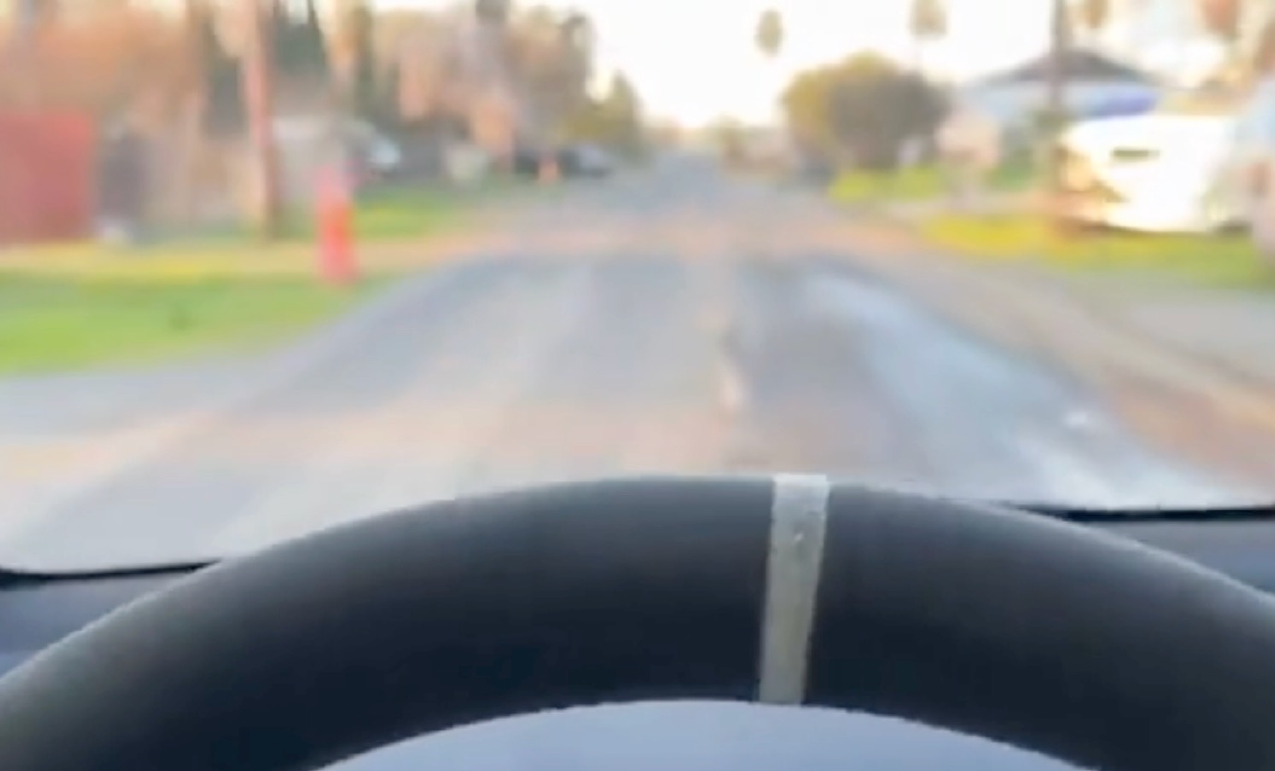}
        \caption{A Tesla autonomous vehicle prioritizes \textit{passenger comfort} over \textit{staying on the correct side}~\cite{elluswamy2025tesla}.}
        \label{fig:tesla_example}
    \end{subfigure}
    \caption{Real-world autonomous driving examples with multiple conflicting objectives.}
    \label{fig:real_world_example}
    \vspace{-4mm}
\end{figure*}

Real-world autonomous driving entails managing multiple, often conflicting objectives with varying priorities. Fig.~\ref{fig:waymo_example} illustrates such a case from the perspective of a Waymo autonomous vehicle~\cite{kxan2024waymo}. A scooter rider in the bike lane suddenly falls onto the road in front of the vehicle. To avoid a collision, the vehicle swerves into the adjacent left lane, nearly crossing into the opposite direction of traffic. In this situation, the vehicle cannot simultaneously satisfy the objectives of \textit{avoiding collision with the rider} and \textit{maintaining lane position}, and therefore prioritizes the former. Fig.~\ref{fig:tesla_example} shows another example~\cite{elluswamy2025tesla}, where a Tesla autonomous vehicle enters the oncoming lane to avoid a large puddle and improve passenger comfort. Here, the objectives of \textit{staying on the correct side of the road} and \textit{ensuring passenger comfort} conflict, and the vehicle prioritizes the latter. These examples highlight the importance of explicitly specifying multiple objectives and their priority relations, along with the environment context, when evaluating autonomous driving systems.

To effectively evaluate autonomous driving systems in modern, complex traffic environments and capture realistic trade-offs among objectives, a benchmark must satisfy three key requirements.
First, it should provide a diverse set of objectives formalized as quantitative metrics and/or Boolean properties for evaluating agents and measuring the degree of violation.
Second, it should include a specification framework that permits expressing priority relations among objectives. This framework must be interpretable, easy to manipulate, and adaptable across scenarios to reflect different preferences.
Third, it should employ an expressive representation of traffic scenarios under which autonomous driving systems can be effectively evaluated under multi-objective specifications, assessing their ability to balance competing goals.

To meet the first requirement, prior work has attempted to formalize traffic rules using Boolean properties or quantitative metrics~\cite{helou2021reasonable, chang2024dynamic, viswanadha2021addressing, shalev2017formal, bellem2016objective, maierhofer2022formalization}. However, these studies do not validate whether the formalized rules accurately capture violations in real driving behaviors. In this paper, we not only formalize a diverse set of driving rules, but also provide alternative definitions for selected rules, serving as a basis for examining how different formulations influence evaluation outcomes. We further design precise and fine-grained measures to assess the degree of rule violation.

Second, most existing autonomous driving benchmarks either focus on a single objective~\cite{ettinger2021large, wilson2023argoverse, caesar2020nuscenes} or consider multiple objectives without modeling their priority relations~\cite{houston2021one, karnchanachari2024towards, caesar2021nuplan, althoff2017commonroad, dauner2024navsim, zhu2025m3cad}. To address this limitation, we adopt the \textit{Rulebook} structure~\cite{censi2019liability}, which captures the priority relations among multiple objectives. We further design a \textit{Hierarchical Rulebook} framework that allows flexible adaptation to diverse driving contexts while remaining interpretable.

Finally, while many driving datasets collect large amounts of human driving data~\cite{ettinger2021large, wilson2023argoverse, caesar2020nuscenes, houston2021one, karnchanachari2024towards, caesar2021nuplan, zhan2019interaction, prabu2022scendd, schuldes2024scenario}, such data-heavy approaches suffer from incompleteness or are inefficient for verification. In our benchmark, we apply the concept of coreset selection~\cite{moser2025coreset, sener2017active} to construct a compact yet representative set of scenarios that achieves broad coverage. We also reconstruct near-accident scenarios from real-world collision reports using a Large Language Model (LLM)-assisted pipeline, enabling evaluation of agents in critical situations. All scenarios are represented in the {\em Scenic programming language}~\cite{fremont2023scenic, fremont2019scenic}, which provides an expressive, yet abstract way to model complex traffic scenarios and stochastically generates diverse concrete scenarios with varying input parameters, enabling parameter-level coverage in simulation-based verification.

The main contributions of this paper are as follows.
\begin{compactItemize}
    \item We propose \textit{ScenicRules}, a benchmark that integrates Scenic programs with Rulebook specifications. To the best of our knowledge, this is the first benchmark to combine a multi-objective, priority-based specification framework (rulebook) with an expressive scenario modeling notation (Scenic) for evaluating autonomous driving approaches and systems.
    \item We collect and formalize a diverse set of autonomous driving objectives with precise, quantitative violation measures (Sec.~\ref{subsec:rule_formalization}).
    \item We design a Hierarchical Rulebook framework that not only encodes multiple prioritized objectives but also remains interpretable, extensible, and adaptable to diverse driving contexts (Sec.~\ref{subsec:our_rulebook}).
    \item We curate a representative and critical set of Scenic scenarios, forming a lightweight yet comprehensive testbed for evaluating autonomous driving systems (Sec.~\ref{subsec:scenarios}).
\end{compactItemize}

In our experiments (Sec.~\ref{sec:experiment}), we validate our formalized objectives and the Hierarchical Rulebook using a public dataset~\cite{helou2021reasonable} containing human trajectory preferences, demonstrating that our formalization aligns with real-world driving behavior. We also analyze the diversity and representativeness of our generated scenarios and evaluate existing driving agents on the scenarios, showing that our benchmark effectively captures agent violations with respect to prioritized objectives.

%% file: sections_arxiv/02_related_work.tex

\section{Related Work}\label{sec:related}
A wide range of autonomous driving benchmarks have been proposed~\cite{liu2024survey}. Many focus on perception and prediction tasks, such as the Waymo Open Motion Dataset~\cite{ettinger2021large}, Argoverse 2~\cite{wilson2023argoverse}, and nuScenes~\cite{caesar2020nuscenes}, which are not for system level evaluation. Our focus is instead on benchmarks for
system-level evaluation, with particular attention to evaluation rules and metrics for driving agents. Lyft Level 5~\cite{houston2021one} and M$^3$CAD~\cite{zhu2025m3cad} employ several metrics, such as collision rate and L2 distance, but do not consider any notion of priority or relative importance among objectives. CommonRoad~\cite{althoff2017commonroad}, nuPlan~\cite{karnchanachari2024towards, caesar2021nuplan}, and NAVSIM~\cite{dauner2024navsim} consider multiple objectives, including collision avoidance, clearance, progress, and comfort, and assign weights to combine them into a single cost function. Competitions such as the CARLA Autonomous Driving Challenge~\cite{carlachallenge} and the NAVSIM Driving Challenge~\cite{dauner2024navsim, cao2025pseudo} similarly aggregate multiple metrics into a single score. Although these weighting schemes encode relative importance to some extent, they still fail to capture explicit priority relations among objectives.

To formally encode priority relations among objectives, the \textit{Rulebook} structure~\cite{censi2019liability} has been proposed. Its detailed definitions are introduced in Sec.~\ref{subsec:rulebook}. The Rulebook framework has been used before in autonomous driving, including applications in minimum-violation planning~\cite{10051644} and unsafe-scenario identification~\cite{collin2020safety}. In~\cite{helou2021reasonable}, a Rulebook instance is designed to classify driving trajectories and compare them with human preferences. Our prior work~\cite{chang2024dynamic,viswanadha2021parallel} developed  falsification algorithms to identify system failures under (small) Rulebook-based specifications, with application to autonomous driving case studies. The present paper goes beyond these works by providing a comprehensive benchmark for evaluation under multi-objective specifications across diverse driving scenarios.
Finally, we note that our approach follows the Verified AI approach~\cite{seshia-cacm22a} combining formal probabilistic modeling of environments with multi-objective formal specifications.
\begin{wrapfigure}{r}{0.25\textwidth}
    \centering
    \includegraphics[width=0.6\linewidth]{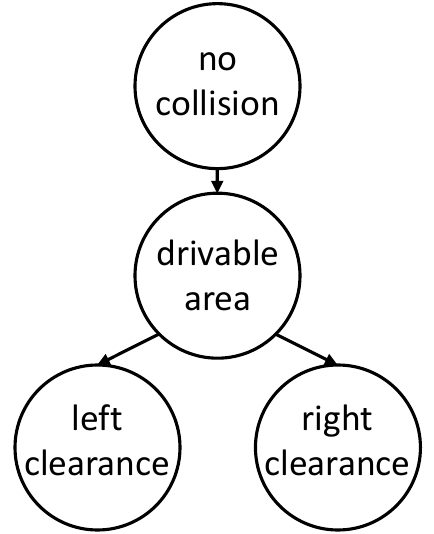}
    \caption{An example of a Rulebook.}
    \label{fig:rulebook_example}
    \vspace{-8mm}
\end{wrapfigure}

%% file: sections_arxiv/03_preliminaries.tex
\section{Preliminaries}\label{sec:prelim}

\subsection{Signal Temporal Logic (STL)} \label{subsec:stl}
Signal temporal logic (STL)~\cite{maler2004monitoring} extends Linear Temporal Logic (LTL) by introducing real-time and real-valued constraints.
In this section, we focus on the two STL operators used in this paper: the \textit{eventually} operator $F$ and the \textit{globally} operator $G$. Their semantics are defined as follows:
\begin{align}
    \nonumber (x, t) \models F_{I\varphi} &\iff \exists t'\in t + I \cdot \big((x, t') \models \varphi\big) \\
    \nonumber (x, t) \models G_{I\varphi} &\iff \forall t'\in t + I \cdot \big((x, t') \models \varphi\big), \label{eq:stl_semantics}
\end{align}
where $x: \mathbb{T} \rightarrow \mathbb{R}^n$ is a signal, $\mathbb{T}$ is the time domain, $I$ is the time interval, and $\varphi$ is an STL formula. The degree of violation of an STL formula can be quantified using the \textit{STL robustness}~\cite{donze2010robust, fainekos2009robustness}. Detailed definitions of STL and its quantitative robustness are provided in Appendix~\ref{appendix:stl}.

\subsection{Rulebook Specification} \label{subsec:rulebook}

The formal definition of the \textit{Rulebook}~\cite{censi2019liability} specification framework is as follows:
\begin{definition}[Rulebook] \label{def:rulebook}
A rulebook $\mathcal{B} = (\mathcal{R}, \preceq_{\mathcal{R}})$ consists of a set of objectives $\mathcal{R}$ and a preorder $\preceq_{\mathcal{R}}$ on $\mathcal{R}$ specifying the priority relations among the objectives. The rulebook can be visualized as a directed graph $\mathcal{G}_\mathcal{B} = \{\mathcal{V}_\mathcal{B}, \mathcal{E}_\mathcal{B}\}$, where each node $v \in \mathcal{V}_\mathcal{B}$ corresponds to an objective, and each directed edge $(v_1, v_2) \in \mathcal{E}_\mathcal{B}$ indicates that $v_1$ has a higher priority than $v_2$. Fig.~\ref{fig:rulebook_example} shows an example.
\end{definition}

Chang et al.~\cite{chang2024dynamic} introduce a metric called \textit{error value} to quantify the degree to which a signal violates a given rulebook.

\begin{definition}[Error Value] \label{def:error_value}
For each objective $r$ in a rulebook, its \textit{error weight} is defined as $2^{m_r}$, where $m_r$ is the number of objectives with lower priority than $r$. Given a signal $x$, the \textit{error value} of $x$ is the sum of the error weights of all objectives violated by $x$. A higher error value indicates a more severe violation of the rulebook.

Using Fig.~\ref{fig:rulebook_example} as an example, the objectives have error weights of $2^3$, $2^2$, $2^0$, and $2^0$ from top to bottom. Suppose a signal $x$ violates \textit{no collision} and \textit{right clearance}, the resulting error value is $2^3 + 2^0 = 9$.
\end{definition}

\subsection{Scenic Programming Language} \label{subsec:scenic}
\begin{figure}[tb]
    \centering
    \includegraphics[width=\linewidth]{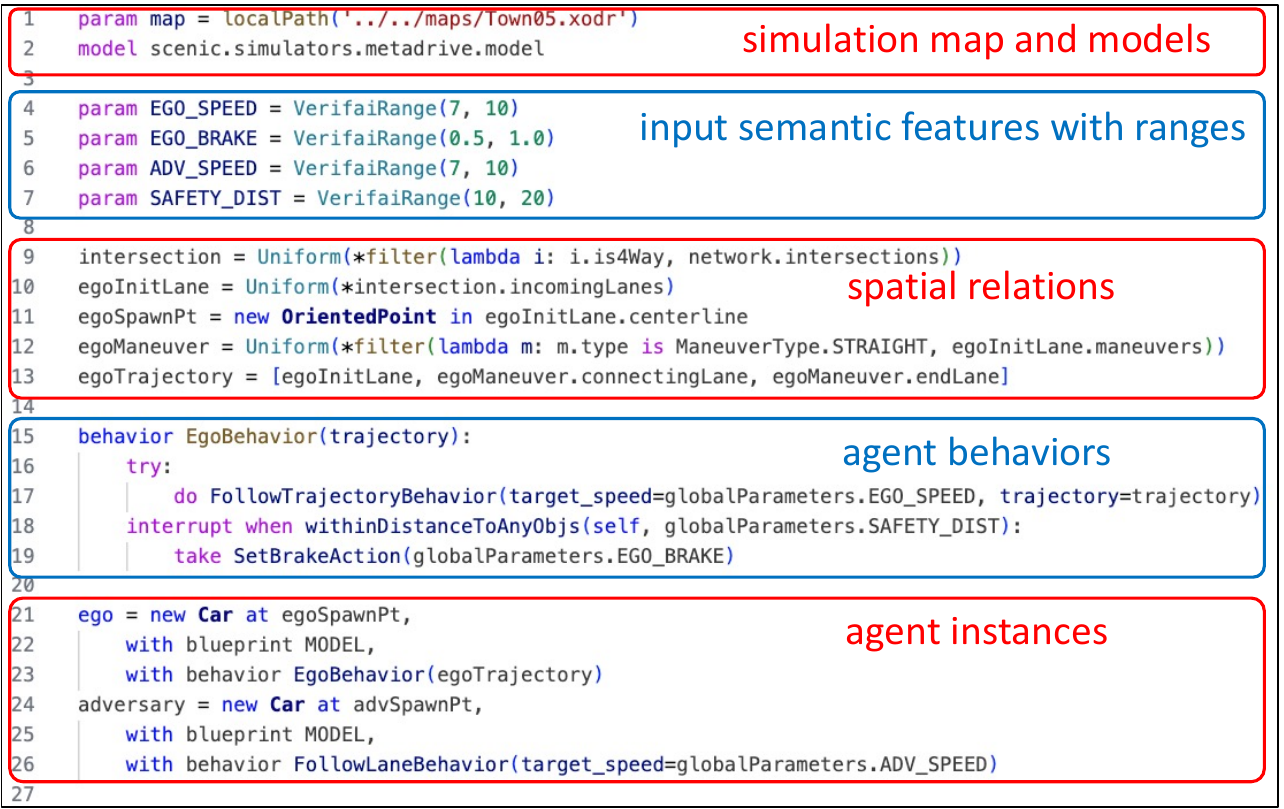}
    \caption{An example Scenic program.}
    \label{fig:scenic_example}
    \vspace{-4mm}
\end{figure}

\textit{Scenic}~\cite{fremont2023scenic, fremont2019scenic} is an open-source domain-specific probabilistic programming system for modeling environments of autonomous systems and robots. Fig.~\ref{fig:scenic_example} shows an example Scenic program for autonomous driving. Scenic provides comprehensive modeling capabilities for road users (e.g., vehicles, pedestrians, cyclists), road network structures, and a library of high-level agent behaviors and low-level control actions. It interfaces with several widely used driving simulators, including CARLA~\cite{dosovitskiy2017carla} and MetaDrive~\cite{li2021metadrive}, enabling seamless simulation and evaluation.

A key strength of Scenic is its ability to specify distributions over regions of semantic parameters and sample/search over them. For instance, in Lines 4–7 of Fig.~\ref{fig:scenic_example}, parameters such as the ego vehicle’s target speed and braking force are defined as ranges. This allows a single Scenic program to represent diverse scenarios under a shared semantic context, offering a compact yet expressive description that is well suited for constructing autonomous driving benchmarks. Furthermore, Scenic supports the evaluation of any component in the autonomous driving stack (e.g., perception, prediction, planning), as well as end-to-end driving systems.

%% file: sections_arxiv/04_benchmark.tex

\section{Our Benchmark}\label{sec:benchmark}

\begin{table*}[t]
    \centering
    \caption{Collected objectives and their formalizations. The notations used in the formulas are defined below the table.}
    \label{tab:objectives}
    \scalebox{0.9}{
    \begin{tabular}{
        |>{\centering\arraybackslash}m{0.02\textwidth}|  
        >{\raggedright\arraybackslash}p{0.13\textwidth}|  
        >{\raggedright\arraybackslash}p{0.36\textwidth}|  
        >{\raggedright\arraybackslash}m{0.43\textwidth}|  
    }
        \hline
        \textbf{Idx} & \textbf{Name} & \textbf{Description} & \textbf{Formula} \\
        \hline\hline
        1 &
        No collisions with VRUs &
        The ego vehicle aims to avoid collisions with vulnerable road users (VRUs), such as pedestrians, cyclists, etc. &
        \makecell[l]{%
            \scriptsize STL: $\operatorname{G}_{[T_1, T_2]}(\forall j\in \mathcal{V} \cdot (p_0(t) \cap p_j(t) = \emptyset))$ \\
            \scriptsize VS: $\sum_{j\in\mathcal{J}} \mathcal{E}_{\text{ego\_loss}} + \mathcal{E}_{\text{vru}_j\text{\_gain}}$, where $\mathcal{E}$ is kinetic energy.
        } \\
        \hline
        2 &
        No collisions with vehicles &
        The ego vehicle aims to avoid collisions with other vehicles. &
        \makecell[l]{%
            \scriptsize STL: $\operatorname{G}_{[T_1, T_2]}(\forall i\in \mathcal{A} \cdot (p_0(t) \cap p_i(t) = \emptyset))$ \\
            \scriptsize VS: $\sum_{i\in\mathcal{A}} \mathcal{E}_{\text{ego\_loss}} + \mathcal{E}_{\text{vehicle}_i\text{\_loss}}$, where $\mathcal{E}$ is kinetic energy.
        } \\
        \hline
        3 &
        Staying within the drivable area &
        The ego vehicle aims to remain within the drivable area. &
        \makecell[l]{
            \scriptsize STL: $\operatorname{G}_{[T_1, T_2]}(p_{0}(t) \subseteq \mathcal{R}_{driv}(t))$ \\
            \scriptsize VS: $\max_t(\|p_{0}(t) \setminus \mathcal{R}_{driv}(t)\| + \operatorname{d}(p_{0}(t), \mathcal{R}_{driv}(t))^2)$
        } \\
        \hline
        4 &
        Staying on the correct side of road &
        The ego vehicle aims to remain on the correct side of road. &
        \makecell[l]{
            \scriptsize STL: $\operatorname{G}_{[T_1, T_2]}(p_{0}(t) \cap (\mathcal{R}_{incorrect}(t) \setminus \mathcal{R}_{correct}(t)) = \emptyset)$ \\
            \scriptsize VS: $\sum_t(\|p_{0}(t) \cap (\mathcal{R}_{incorrect}(t) \setminus \mathcal{R}_{correct}(t))\|)$
        } \\
        \hline
        5 &
        VRU time to collision &
        The ego aims to maintain a time to collision (TTC) above a threshold for all VRUs, assuming the ego maintains its current velocity while VRUs stop suddenly. &
        \makecell[l]{
            \scriptsize STL: $\operatorname{G}_{[T_1, T_2]}(\forall j \in \mathcal{V} \cdot (TTC_j(t) \geq T_{VRU,ttc}))$ \\
            \scriptsize VS: $\max_{t, j} (T_{VRU,ttc} - TTC_j(t))$
        } \\
        \hline
        6 &
        VRU acknowledgment &
        If the ego is expected to come within a minimum safe distance of a VRU and its current velocity towards the VRU is higher than a threshold, it aims to decelerate sufficiently. &
        \makecell[l]{
            \scriptsize STL: $\operatorname{G}_{[T_1, T_2]}(\forall \tau \in [t, t+T_{ack}] \cdot ((\exists j \in \mathcal{V} \cdot ((v_{0,j}(t) > V_{ack})$ \\ 
            \scriptsize $\land (\operatorname{d}(p_0(\tau), p_j(\tau)) < D_{ack})) \implies a_{0, j}(t) < A_{ack}))$ \\
            \scriptsize VS: $\max_{t_k, j} (a_{0, j}(t) - A_{ack})$
        } \\
        \hline
        7 &
        Vehicle time to collision &
        The ego aims to maintain a time to collision (TTC) above a threshold for any other vehicle, assuming both vehicles maintain their current velocities. &
        \makecell[l]{
            \scriptsize STL: $\operatorname{G}_{[T_1, T_2]}(\forall i \in \mathcal{A} \cdot (TTC_i(t) \geq T_{vehicle,ttc}))$ \\
            \scriptsize VS: $\max_{t_k, i} (a_{0, i}(t) - A_{ack})$
        } \\
        \hline
        8 &
        VRU off-road clearance &
        The ego vehicle aims to maintain a distance greater than a specified threshold from all off-road VRUs. &
        \makecell[l]{
            \scriptsize STL: $\operatorname{G}_{[T_1, T_2]}\big(\forall j \in \mathcal{V}_{off}(t) \cdot (\operatorname{d}(p_0(t), p_j(t)) > D_{VRU, off})\big)$ \\
            \scriptsize VS: $\max_{t, j \in \mathcal{V}_{off}(t)} \big(D_{VRU, off} - \operatorname{d}(p_0(t), p_j(t))\big)$
        } \\
        \hline
        9 &
        VRU on-road clearance &
        The ego vehicle aims to maintain a distance greater than a specified threshold from all on-road VRUs. &
        \makecell[l]{
            \scriptsize STL: $\operatorname{G}_{[T_1, T_2]}\big(\forall j \in \mathcal{V}_{on}(t) \cdot (\operatorname{d}(p_0(t), p_j(t)) > D_{VRU, on})\big)$ \\
            \scriptsize VS: $\max_{t, j \in \mathcal{V}_{on}(t)} \big(D_{VRU, on} - \operatorname{d}(p_0(t), p_j(t))\big)$
        } \\
        \hline
        10 &
        Front vehicle clearance &
        The ego vehicle aims to maintain a distance greater than a specified threshold from all vehicles ahead. &
        \makecell[l]{
            \scriptsize STL: $\operatorname{G}_{[T_1, T_2]}\big(\forall i \in \mathcal{A}_{front}(t) \cdot (\operatorname{d}(p_0(t), p_i(t)) > D_{vehicle, front})\big)$ \\
            \scriptsize VS: $\max_{t, i \in \mathcal{A}_{front}(t)} \big(D_{vehicle, front} - \operatorname{d}(p_0(t), p_i(t))\big)$
        } \\
        \hline
        11 &
        Left vehicle clearance &
        The ego vehicle aims to maintain a distance greater than a specified threshold from all vehicles on the left. &
        \makecell[l]{
            \scriptsize STL: $\operatorname{G}_{[T_1, T_2]}\big(\forall i \in \mathcal{A}_{left}(t) \cdot (\operatorname{d}(p_0(t), p_i(t)) > D_{vehicle, left})\big)$ \\
            \scriptsize VS: $\max_{t, i \in \mathcal{A}_{left}(t)} \big(D_{vehicle, left} - \operatorname{d}(p_0(t), p_i(t))\big)$
        } \\
        \hline
        12 &
        Right vehicle clearance &
        The ego vehicle aims to maintain a distance greater than a specified threshold from all vehicles on the right. &
        \makecell[l]{
            \scriptsize STL: $\operatorname{G}_{[T_1, T_2]}\big(\forall i \in \mathcal{A}_{right}(t) \cdot (\operatorname{d}(p_0(t), p_i(t)) > D_{vehicle, right})\big)$ \\
            \scriptsize VS: $\max_{t, i \in \mathcal{A}_{right}(t)} \big(D_{vehicle, right} - \operatorname{d}(p_0(t), p_i(t))\big)$
        } \\
        \hline
        13 &
        Speed limit &
        The ego vehicle aims to maintain a speed below the posted speed limit. &
        \makecell[l]{
            \scriptsize STL: $\operatorname{G}_{[T_1, T_2]}\big(\|v_0(t)\| < V_{limit} \big)$ \\
            \scriptsize VS: $\big(\max\big(\max_{t}(\|v_0(t)\| - V_{limit}), 0\big)\big)^2$
        } \\
        \hline
        14 &
        Lane keeping &
        The ego vehicle aims to remain within its current lane. &
            \scriptsize Obj. func.: $\min \big( \Sigma_{t=T_1}^{T_2} \mathds{1}_{(p_0(t) \nsubseteq \mathcal{R}_{0, lane}(t))} (t) \big)$
        \\
        \hline
        15 &
        Lane centering &
        The ego vehicle aims to stay close to the centerline of its lane. &
        \makecell[l]{
            \scriptsize Obj. func.: $\min \big(\Sigma_{t=T_1}^{T_2} \operatorname{d}(c_0(t), \mathcal{L}_{centerline}(t)) \big)$, where $c_0(t)$ is \\ 
            \scriptsize the ego centroid, $\mathcal{L}_{centerline}(t)$ is the centerline of the current lane.
        } \\
        \hline
        16 &
        Goal / Progress &
        The ego vehicle aims to make progress and eventually reach the target region. &
        \makecell[l]{
            \scriptsize STL: $\operatorname{F}_{[T_1, T_2]}\big(p_0(t) \subseteq \mathcal{R}_{target}(t) \big)$ \\
            \scriptsize VS: $\min_t(\|p_{0}(t) \setminus \mathcal{R}_{target}(t)\| + \operatorname{d}(p_{0}(t), \mathcal{R}_{target}(t))^2)$
        } \\
        \hline
        17--19 &
        Comfort &
        The ego vehicle aims to minimize jerk (the rate of change of acceleration), longitudinal, and lateral acceleration. &
        \makecell[l]{
            \scriptsize Obj. funcs.: $\min \big( \Sigma_{t=T_1+1}^{T_2} \|a_0(t)-a_0(t-1)\| \big)$, \\
            \scriptsize $\min \big( \Sigma_{t=T_1}^{T_2} \left|a_{0, long}(t)\right| \big)$,
            \scriptsize $\min \big( \Sigma_{t=T_1}^{T_2} \left|a_{0, lat}(t)\right| \big)$
        } \\
        \hline
    \end{tabular}
    }
    \begin{tablenotes}
        \item Notations in the formulas: $[T_1, T_2]$ is the simulation time interval. $0$ denotes the ego index, $i \in \mathcal{A}$ other vehicles, and $j \in \mathcal{V}$ vulnerable road users (VRUs). $p_{0}(t), p_{i}(t), p_{j}(t)$ are the polygons of the ego, vehicle $i$, and VRU $j$, respectively. $\mathcal{R}_{driv}(t), \mathcal{R}_{correct}(t), \mathcal{R}_{incorrect}(t), \text{and } \mathcal{R}_{target}(t)$ denote the drivable area, correct side of the road, incorrect side of the road, and target region. $t_k$ are the time steps where a violation occurs. Variables prefixed by $T, D, V,$ and $A$ represent thresholds for time, distance, velocity, and acceleration, respectively. $\operatorname{d}(\cdot, \cdot)$ returns the minimum distance between two objects, and $\|\cdot\|$ denotes the polygon area or vector magnitude.
    \end{tablenotes}
    \vspace{-4mm}
\end{table*}

In this section, we present the step-by-step process of constructing our multi-objective benchmark. In Sec.~\ref{subsec:rule_formalization}, we collect 19 autonomous driving objectives from the literature and formalize them. In Sec.~\ref{subsec:our_rulebook}, we introduce the Rulebook specification structure and propose our hierarchical rulebook to systematically encode objective priorities. Finally, in Sec.~\ref{subsec:scenarios}, we create representative and critical scenarios as the testbed for evaluating autonomous driving agents against our multi-objective specifications.

\subsection{Collection and Formalization of Objectives}\label{subsec:rule_formalization}

We begin by collecting a diverse set of autonomous driving objectives from prior works~\cite{helou2021reasonable, chang2024dynamic, viswanadha2021addressing, shalev2017formal, bellem2016objective}, as detailed in Table~\ref{tab:objectives}. Although not exhaustive, this represents the most extensive collection of formalized rules to date and is designed for future extensibility.

After collection, we formalize each objective to enable systematic evaluation of autonomous driving agents. Once formalized, we refer to these objectives as \textit{rules} throughout the paper. The formalization must (i) determine whether an agent satisfies or violates a rule and (ii) quantify the extent of violation. \textit{Signal Temporal Logic (STL)}~\cite{maler2004monitoring} is a convenient option that supports both capabilities, though our approach is not restricted to STL. Each rule is defined by a logical formula together with a corresponding \textit{violation score (VS)} that measures the extent of violation. A positive VS indicates a violation, and larger values correspond to more severe violations. Below, we highlight three representative examples:

\textbf{1) Rule 3 (staying within the drivable area):} The STL formula enforces that the ego vehicle’s polygon remains entirely within the drivable area throughout the simulation. The corresponding VS inherits the STL robustness metric by using the out-of-bound area to measure violation severity (first term). However, once the ego vehicle is entirely outside the drivable region, the out-of-bound area would remain constant regardless of how far the vehicle continues to move away from the drivable region. Thus, we introduce a distance-based term (second term) to further differentiate degrees of violation. This example illustrates how we refine STL robustness for more precise quantification.

\textbf{2) Rule 10 -- 12 (vehicle clearance):} The default violation score (VS) for these rules is defined as the maximum clearance violation along the trajectory. We also consider an alternative formalization in which the VS is computed as the sum of all clearance violations. This distinction can change which trajectories are preferred, since the default favors avoiding any single extreme violation, whereas the alternative penalizes trajectories with continual small violations. This example demonstrates that different rule formalizations can yield different evaluation outcomes.

\textbf{3) Rule 15 (lane centering):} This example shows that rules need not be limited to STL formulations. Here, we adopt the \textit{objective function} to define the rule, aiming to minimize the ego's overall deviation from the centerline. Note that this can be easily converted into an STL formula by introducing a deviation threshold and enforcing that the deviation remains below it over time.

In our implementation, each rule is defined as a Python function that takes the simulation results as input and outputs the VS. Users can easily extend our framework by adding custom rules. Furthermore, our framework supports {\em both} STL-based and objective-function-based formulations, enabling flexible and scalable definition of evaluation metrics.

\subsection{Hierarchical Rulebook Specifications}\label{subsec:our_rulebook}
\begin{figure}[tb]
    \centering
    \includegraphics[width=\linewidth]{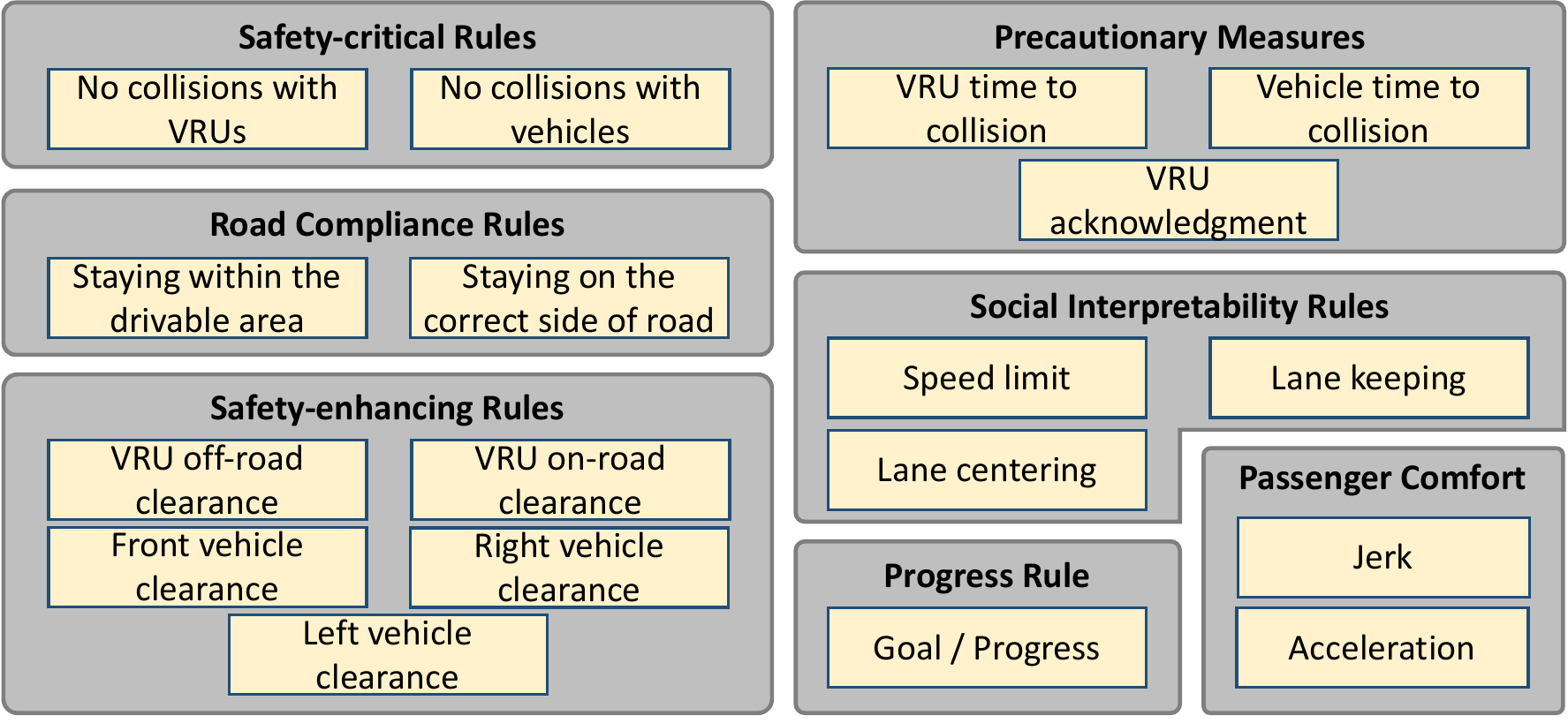}
    \caption{Categorization of rules.}
    \label{fig:rule_categorization}
    \vspace{-2mm}
    \caption*{\footnotesize 
    \textbf{Safety-Critical Rules}: Rules directly impacting the safety of road users. \\
    \textbf{Road Compliance Rules}: Rules enforcing adherence to road structures. \\
    \textbf{Safety-Enhancing Rules}: Rules that may not correspond to immediate safety violations but promote safer and smoother traffic flow when followed. \\
    \textbf{Precautionary Measures}: Rules that help mitigate potential safety risks in foreseeable future conditions. \\
    \textbf{Social Interpretability Rules}: Rules that improve the ego vehicle’s predictability and reduce uncertainty for surrounding road users. \\
    \textbf{Progress Rule}: Ensures the ego vehicle makes progress toward its destination. \\
    \textbf{Passenger Comfort Rules}: Rules promoting smoothness and comfort for passengers. \\
    }
    \vspace{-4mm}
\end{figure}

After collecting and formalizing the rules, we adopt the \textit{Rulebook} framework~\cite{censi2019liability} to encode specifications involving multiple objectives and their priority relations. As discussed in the introduction, different driving scenarios may require different Rulebook specifications to reflect varying priority preferences. However, directly redefining the pairwise priorities among all rules is highly inefficient. Among the 19 collected rules, there exist more than $10^{35}$ possible orders~\cite{sloane}, most of which are difficult to interpret or justify. Therefore, a Rulebook structure that remains interpretable and can be easily adapted to diverse scenarios is desirable.

To address this challenge, we first observe that rules of similar types often share similar priorities. For example, left and right vehicle clearances are typically of equal importance in most driving contexts. This suggests that rules can be grouped based on functional similarity, assigning comparable priorities within each group.
Moreover, relations between groups are generally hierarchical: if one group outranks another (e.g., collision avoidance over passenger comfort), all its constituent rules take precedence.
Building on these observations, we propose the concept of a \textit{Hierarchical Rulebook}\footnote{In an earlier version, this was referred to as a ``Modularized Rulebook''.}.

\begin{definition}[Hierarchical Rulebook]\label{def:modularized_rulebook}
A Hierarchical Rulebook $\mathcal{B}_M$ is a tuple $(\mathcal{R}_M, \preceq_{\mathcal{R}_M})$, where $\mathcal{R}_M$ is a set of rule groups, and each group $\mathcal{S} \in \mathcal{R}_M$ consisting of one or more rules. The preorder $\preceq_{\mathcal{R}_M}$ defines 
inter-group priorities: if group $\mathcal{S}_1$ has higher priority than $\mathcal{S}_2$, all rules in $\mathcal{S}_1$ take precedence over those in $\mathcal{S}_2$. Each group functions as an internal rulebook, with intra-group priorities determined by predefined principles (e.g., equal rank or higher criticality for VRU-related rules).
\end{definition}

The Hierarchical Rulebook allows users to adjust priorities at the group level rather than at the individual rule level, significantly simplifying customization and enhancing interpretability. The categorization of rules and the underlying rationale for each group are illustrated in Fig.~\ref{fig:rule_categorization}. It is important to note that because different driving contexts necessitate distinct priority relations, any group orderings presented in this paper serve only as reference baselines. Users can freely adapt these relations to fit specific environments. The following examples demonstrate how group-level priorities can be adjusted to handle diverse scenarios.
\begin{example}[Emergency Vehicles]
If the ego vehicle is an emergency vehicle on duty, the \textit{progress rule} may be prioritized over \textit{road compliance rules} (e.g., an ambulance may temporarily drive against the flow of traffic).
\end{example}
\begin{example}[Dense Traffic]
In dense, complex traffic such as busy intersections, hesitation or ambiguous behavior can cause confusion and safety risks. In this case, \textit{social interpretability rules} should be prioritized to ensure clarity of intent.
\end{example}
\begin{example}[Degraded Performance]
If the ego vehicle experiences degraded braking performance, it must maintain greater distance from other agents. Hence, \textit{precautionary measures} and \textit{safety-enhancing rules} become more critical.
\end{example}

These examples demonstrate how group-level priority relations can be permuted to accommodate different driving contexts. To enable systematic and automated adaptation, we further propose a data-driven priority optimization framework. Given a default hierarchical rulebook and a set of trajectories with annotated human preferences, our algorithm iteratively refines the rulebook by greedily swapping priority relations between rule groups to maximize alignment with the provided annotations. This process continues until convergence, allowing users to automatically derive a rulebook that reflects their desired preferences without manual design. Detailed mechanics of this algorithm are provided in Appendix~\ref{appendix:priority}. However, it is worth noting that accurately learning preferences directly from data remains a non-trivial challenge, representing a direction for future work.

\subsection{Formalized Environment Scenarios in Scenic}\label{subsec:scenarios}

After constructing the hierarchical rulebook specifications, we proceed to create representative and critical scenarios. We adopt Scenic~\cite{fremont2023scenic, fremont2019scenic} as the scenario description language for its ability to express probabilistic variations at the parameter level and its comprehensive modeling of autonomous driving environments (as discussed in Sec.~\ref{subsec:scenic}). Our benchmark includes two types of scenarios: \textit{common scenarios} and \textit{near-accident scenarios}. The common scenarios test typical driving maneuvers and spatial relations, while the near-accident scenarios focus on rare, safety-critical cases to evaluate agents’ responses in high-risk situations.

\subsubsection{Common Scenarios}\label{subsubsec:common_scenarios}
\begin{figure}
    \centering
    \includegraphics[width=0.9\linewidth]{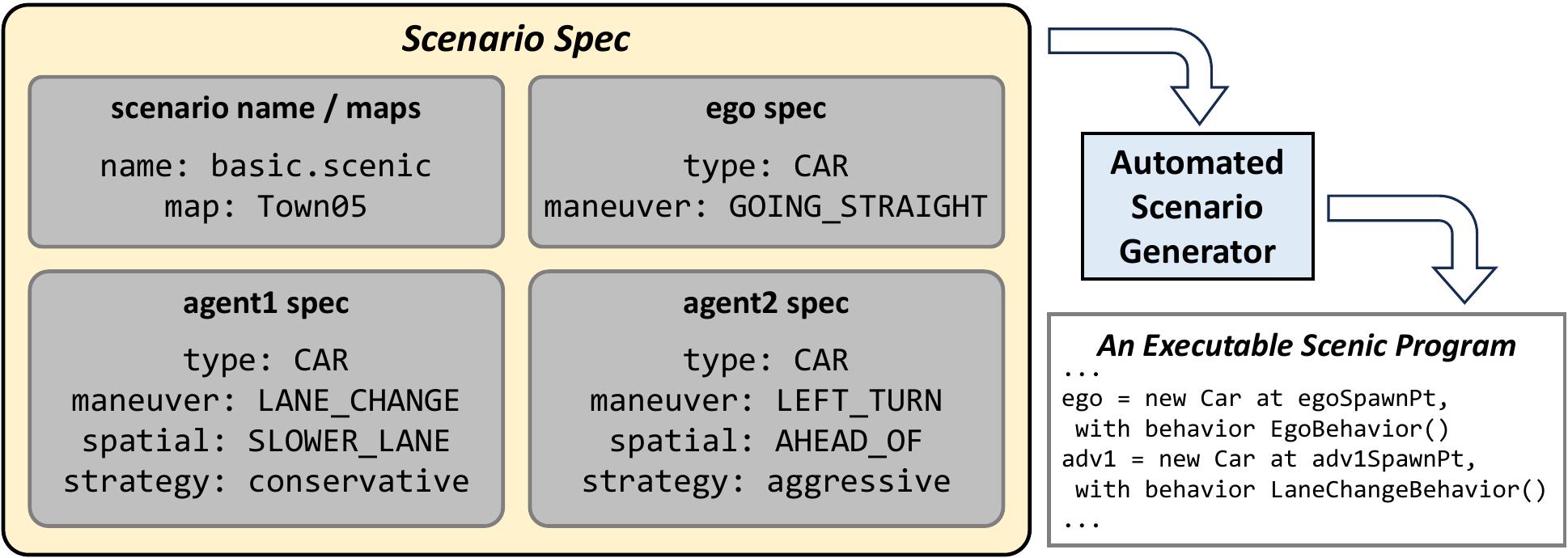}
    \caption{Our automated Scenic program generator.}
    \label{fig:scenario_generator}
    \vspace{-2mm}
\end{figure}

\begin{table}[tb]
    \centering
    \caption{Supported attributes in the Scenic generator.}
    \label{tab:scenario_generator_options}
    \scalebox{0.9}{
    \begin{tabular}{|p{0.3\linewidth}|p{0.6\linewidth}|}
    \hline
    \textbf{Attribute} & \textbf{Supported Options} \\
    \hline\hline
    agent type & car, pedestrian \\
    \hline
    vehicle maneuver & going straight, left turn, right turn, lane change, lane following \\
    \hline
    spatial relation to ego & ahead of, behind, faster lane, slower lane, opposing lanes, conflicting lanes \\
    \hline
    pedestrian maneuver & crossing street, walking along sidewalk \\
    \hline
    \end{tabular}
    }
    \vspace{-2mm}
\end{table}

To systematically evaluate typical maneuvers and spatial relations in traffic, we develop an automated Scenic program generator (Fig.~\ref{fig:scenario_generator}). Users specify the map, ego maneuver, and other agents’ attributes (types, maneuvers, spatial relations) in a JSON-style format, and the generator outputs an executable Scenic program. Table~\ref{tab:scenario_generator_options} lists all supported options, and Appendix~\ref{appendix:common_scenario} provides a complete example.

This generator defines a vast space of possible programs. For example, with up to three surrounding agents, there are over one million possible combinations of agent types, maneuvers, and spatial relations. 
Since testing every combination is infeasible, a key question arises:
\textit{Which scenarios should be included in the benchmark?} Ideally, the selected subset should be both \textit{diverse} and \textit{representative}, covering a broad range of agent behaviors while minimizing redundancy.

To achieve this, we draw inspiration from 
\textit{coreset selection} in machine learning~\cite{moser2025coreset}. Given a large dataset $T$, coreset selection seeks a small subset $S$ ($|S| \ll |T|$) such that a model trained on $S$ performs comparably to one trained on $T$. Analogously, we aim to find a compact set of scenarios that captures the diversity of the entire scenario space while preserving evaluation effectiveness.

A common approach for coreset selection is the \textit{k-Center Greedy} algorithm~\cite{sener2017active}, which minimizes the maximum distance between any unselected point $x \in T \setminus S$ and its nearest selected point $x' \in S$. In each iteration, it selects the point with the largest minimum distance to the current subset, repeating until the desired number of samples is reached. To apply this to scenario selection, each Scenic program is encoded as a vector, where each dimension represents a discrete attribute of the scenario. The first element encodes the ego’s maneuver, and each subsequent pair of elements encodes the spatial relation and maneuver of another agent.

\begin{example}[Scenario Encoding]
Assume integers represent vehicle maneuvers (e.g., going straight (1), left turn (2), right turn (3)) and letters represent spatial relations (e.g., ahead (A), behind (B), faster lane (C)). A vector $[3, A, 1, C, 2]$ encodes an ego vehicle performing a right turn (encoded as 3), one agent ahead (A) going straight (1), and another agent in the faster lane (C) turning left (2).
\end{example}

Using this encoding, we use the Hamming distance~\cite{hamming1950error} to quantify the difference between any two scenarios (i.e., the number of different elements in their corresponding vectors). Since the order of non-ego agents is interchangeable, we compute Hamming distances across all permutations and take the minimum value. Then, applying the k-Center Greedy algorithm, we iteratively select the scenario with the largest minimum distance to the already selected ones. This ensures that every scenario in the original Scenic program space is close to at least one selected scenario, guaranteeing representativeness while covering diverse combinations of maneuvers and spatial relations.

\subsubsection{Near-accident Scenarios}\label{subsubsec:near_accident_scenarios}
While the common scenarios provide broad coverage of typical driving situations, they do not capture extreme cases--such as a leading vehicle abruptly braking with maximum force or situations resembling Fig.~\ref{fig:waymo_example}. To address this gap, we encode into Scenic near-accident scenarios from autonomous vehicle collision reports published by the California Department of Motor Vehicles (DMV)~\cite{californiadmv2025report}.

Following a similar setup to ScenicNL~\cite{elmaaroufi2024scenicnl}, each collision report is summarized into a one-paragraph natural language description. Unlike ScenicNL, which employs a compositional prompting strategy with iterative feedback, we adopt a simpler few-shot prompting method. As shown in Fig.~\ref{fig:scenic_example}, 
Scenic programs follow a distinct structure; our prompt explicitly defines this schema and outlines essential syntax elements (e.g., built-in agent behaviors) to guide generation, supplemented by a few $\langle$natural language, Scenic program$\rangle$ example pairs.
Using the Gemini 2.5 Flash~\cite{comanici2025gemini} model, each Scenic program can be generated in about 30 seconds, typically requiring only minor human refinement. The complete prompt is provided in Appendix~\ref{appendix:llm}.

%% file: sections_arxiv/05_experiments.tex
\section{Experiments}\label{sec:experiment}

We aim to address two main research questions:
\begin{compactItemize}
    \item \textbf{RQ1:} Do our formalized rules and proposed hierarchical rulebook structure accurately capture the objectives and priority preferences observed in real-world driving behaviors?
    \item \textbf{RQ2:} Can the scenarios in our benchmark effectively and efficiently evaluate autonomous driving agents with respect to the prioritized objectives?
\end{compactItemize}

\begin{table}[t]
    \centering
    \caption{Evaluation results against human preferences.}
    \label{tab:reasonable_crowd_results}
    \scalebox{0.9}{
        \begin{tabular}{|c|c|}
        \hline
        \textbf{Model} & \textbf{Accuracy (\%)} \\
        \hline\hline
        Ours & $80.9 \pm 1.8$ \\
        \hline
        RB + DT & $78.4 \pm 2.8$ \\
        \hline
        RF & $82.3 \pm 2.4$ \\
        \hline
        LR & $79.1 \pm 2.5$ \\
        \hline\hline
        w/ parameter & $81.8 \pm 2.0$ \\
        \hline
        \end{tabular}
    }
    \begin{tablenotes}
    \item RB + DT, RF, and LR denote the rulebook + decision tree, random forest, and logistic regression models from~\cite{helou2021reasonable}, respectively. 
    \item ``Ours'' is the base hierarchical rulebook; ``w/ parameter'' indicates results after rule-parameter optimization.
    \end{tablenotes}
    \vspace{-1.5mm}
\end{table}

\begin{table}
    \centering
    \caption{Scenario-specific priority optimization results.}
    \label{tab:scenario_rulebook_results}
    \scalebox{0.9}{
        \begin{tabular}{|c|c|c|}
        \hline
        \textbf{Restricted} & \textbf{\# Iterations} & \textbf{Accuracy (\%)} \\
        \hline\hline
        Yes & $2$ & $85.08$ \\
        \hline
        No & $2$ & $85.97$ \\
        \hline
        \multicolumn{2}{|c|}{Brute Force Search} & $85.97$ \\
        \hline
        \end{tabular}
    }
    \begin{tablenotes}
    \item ``Restricted'' indicates that only adjacent groups can be swapped. ``\#~Iterations'' denotes the number of greedy iterations until convergence. ``Brute Force Search'' reports the best accuracy among all possible priority permutations.
    \end{tablenotes}
    \vspace{-1.5mm}
\end{table}

\begin{table}[t]
\centering
    \caption{Evaluation results under alternative rule formalizations.}
    \label{tab:alternative_rule_results}
    \scalebox{0.9}{
        \begin{tabular}{|c|ccc|}
        \hline
        \textbf{Formalization} &
          \textbf{Accuracy (\%)} &
          \textbf{\begin{tabular}[c]{@{}c@{}}\# Different\\ Preferences\end{tabular}} &
          \textbf{\begin{tabular}[c]{@{}c@{}}\# Reason\\ Changes\end{tabular}} \\ \hline\hline
        Clearance -- Sum                                                   & $80.92$ & $6$  & $1$   \\ \hline
        Clearance -- Heading                                               & $81.63$ & $23$  & $251$   \\ \hline
        Correct Side -- Centroid                                           & $78.95$ & $69$   & $171$   \\ \hline\hline
        Original                                                           & $80.97$ & --     & --      \\ \hline
        \end{tabular}
    }
    \begin{tablenotes}
        \item \textit{Clearance -- Sum}: As described in the vehicle clearance example in Sec.~\ref{subsec:rule_formalization}.
        \item \textit{Clearance -- Heading}: Instead of using the ego’s future trajectory to categorize front, left, and right vehicles (default), this variant classifies them based on their position relative to ego heading.
        \item \textit{Correct Side -- Centroid}: Rather than requiring the entire ego polygon to lie on the correct side (default), this variant checks only whether the ego’s centroid is on the correct side.
        \item ``\# Reason Changes'' counts how many times the key rule determining the final preference differs from that of the original rulebook. 
        \item The experiments for this table are conducted on the whole dataset (not 5-fold split), hence the original accuracy is slightly different from that in Table~\ref{tab:reasonable_crowd_results}.
    \end{tablenotes}
    \vspace{-5mm}
\end{table}

\subsection{RQ1: Evaluation of Hierarchical Rulebooks}\label{subsec:rq1}

To evaluate the alignment of our formalized rules and hierarchical rulebook with human driving preferences, we utilize the \textit{Reasonable Crowd} dataset~\cite{helou2021reasonable}. This dataset contains 92 scenarios, each containing several feasible trajectories with pairwise human preference labels. 
For each trajectory pair, we compare the preference induced by our hierarchical rulebook against the human-derived preference. Detailed descriptions of the dataset and the evaluation pipeline are provided in Appendix~\ref{appendix:eval_human}. We adopt the following baseline priority ordering: safety-critical $\succ$ road-compliance $\succ$ safety-enhancing $\succ$ social interpretability $\succ$ precautionary measures. Progress and passenger-comfort rules are excluded due to dataset limitations; however, as they hold the lowest priority, their inclusion would not degrade performance.

\paragraph{Performance and Generalizability}
Table~\ref{tab:reasonable_crowd_results} summarizes the evaluation results. Our base rulebook achieves accuracy comparable to the RF and LR baselines from~\cite{helou2021reasonable}. While these learning-based methods rely on training data, our rulebook is designed independently of the dataset and generalizes to arbitrary scenarios.
Furthermore, our hierarchical structure is inherently more interpretable than the learning-based models and also outperforms the rulebook instance from~\cite{helou2021reasonable}. Regarding the approximately $20\%$ discrepancy between our rulebook and human preferences, we attribute a significant portion to inherent variability in human judgment. As reported in~\cite{helou2021reasonable}, the median agreement rate between individual annotators and the aggregate preference is roughly $84\%$, suggesting that human preferences diverge in identical driving contexts; hence, perfect alignment between a rulebook and human consensus across all trajectory pairs is inherently unlikely.

\paragraph{Rule Parameter Optimization}
Since parameters in rule formalizations influence the evaluation outcomes (e.g., a larger clearance threshold increases the likelihood of rule violations), we perform parameter optimization to determine suitable values. 
As higher-priority rules dominate preference comparisons, we traverse the rule hierarchy in topological order, greedily selecting parameter values that maximize performance. The detailed optimization algorithm is provided in Appendix~\ref{appendix:parameter}.
To ensure a fair comparison, we tune parameters using the same 5-fold split and 10 random seeds employed by the baselines in~\cite{helou2021reasonable}. As shown in Table~\ref{tab:reasonable_crowd_results}, this optimization further improves accuracy.

\paragraph{Context-Aware Priority Tuning}
As discussed in Sec.~\ref{subsec:our_rulebook}, optimal priority relations between rule groups are often context-dependent. Therefore, we conduct priority tuning to build a scenario-specific rulebook for each scenario in the Reasonable Crowd dataset. Using the greedy algorithm from Sec.~\ref{subsec:our_rulebook}, and with the safety-critical group fixed at the highest priority, we optimize the priority relations among the remaining groups. We evaluate two settings: swapping only adjacent groups and swapping arbitrary pairs. As shown in Table~\ref{tab:scenario_rulebook_results}, scenario-specific rulebooks achieve higher accuracy than a monolithic rulebook, validating that human preferences are context-dependent. Moreover, the greedy algorithm matches brute-force accuracy and converges within two iterations, highlighting its efficiency.

We illustrate this adaptation with a specific scenario where multiple pedestrians closely approach the ego vehicle. The default rulebook favors evasive maneuvers (e.g., lane changes) to maximize clearance, prioritizing safety-enhancing rules over social interpretability. However, human annotators largely prefer the ego vehicle to maintain its course rather than execute abrupt maneuvers, as the clearance violation is caused by the pedestrians. Consequently, the optimized rulebook for this scenario prioritizes social interpretability over safety-enhancing rules.
This case also suggests that rule rankings are influenced by the attribution of responsibility; distinguishing between ego- and adversary-initiated violations could prevent such misclassifications. Our priority optimization flow could identify these instances for future refinement.

\paragraph{Impact of Rule Formalization}
We compare evaluation outcomes under three alternative rule formalizations (Table~\ref{tab:alternative_rule_results}). Modifying these formalizations leads to noticeable changes in the rulebook’s selected preferences, affecting both the number of preference deviations and shifts in the determining rules (``reason''). These findings highlight that precise rule formalization significantly impacts evaluations, underscoring the importance of contextually appropriate rule definitions.

\subsection{RQ2: Evaluation of the Scenarios}\label{subsec:rq2}

\begin{table}[t]
    \centering
    \caption{Scenario coverage of common scenarios.}
    \label{tab:scenario_coverage_results}
    \scalebox{0.85}{
        \begin{tabular}{|ccc|ccccc|}
        \hline
          \textbf{Config.} &
          \textbf{\begin{tabular}[c]{@{}c@{}}Space\\ Size\end{tabular}} &
          \textbf{Dim.} &
          \textbf{\begin{tabular}[c]{@{}c@{}}Max.\\ Dist.\end{tabular}} &
          \textbf{\begin{tabular}[c]{@{}c@{}}Ego\\ Man.\end{tabular}} &
          \textbf{\begin{tabular}[c]{@{}c@{}}Adv.\\ Man.\end{tabular}} &
          \textbf{\begin{tabular}[c]{@{}c@{}}Adv.\\ Spat.\end{tabular}} &
          \textbf{\begin{tabular}[c]{@{}c@{}}Spat.-\\ Man.\end{tabular}} \\ \hline\hline
        \begin{tabular}[c]{@{}c@{}}2 adversaries\\ no pedestrian\end{tabular} &
          $1616 $ &
          $5    $ &
          $2    $ &
          $100\%$ &
          $100\%$ &
          $100\%$ &
          $100\%$ \\ \hline
        \begin{tabular}[c]{@{}c@{}}2 adversaries\\ 1 pedestrian\end{tabular} &
          $3232 $ &
          $7    $ &
          $2    $ &
          $100\%$ &
          $100\%$ &
          $100\%$ &
          $100\%$ \\ \hline
        \end{tabular}
    }
    \vspace{-1mm}
\end{table}

\begin{figure}[t]
    \centering
    \includegraphics[width=0.9\linewidth]{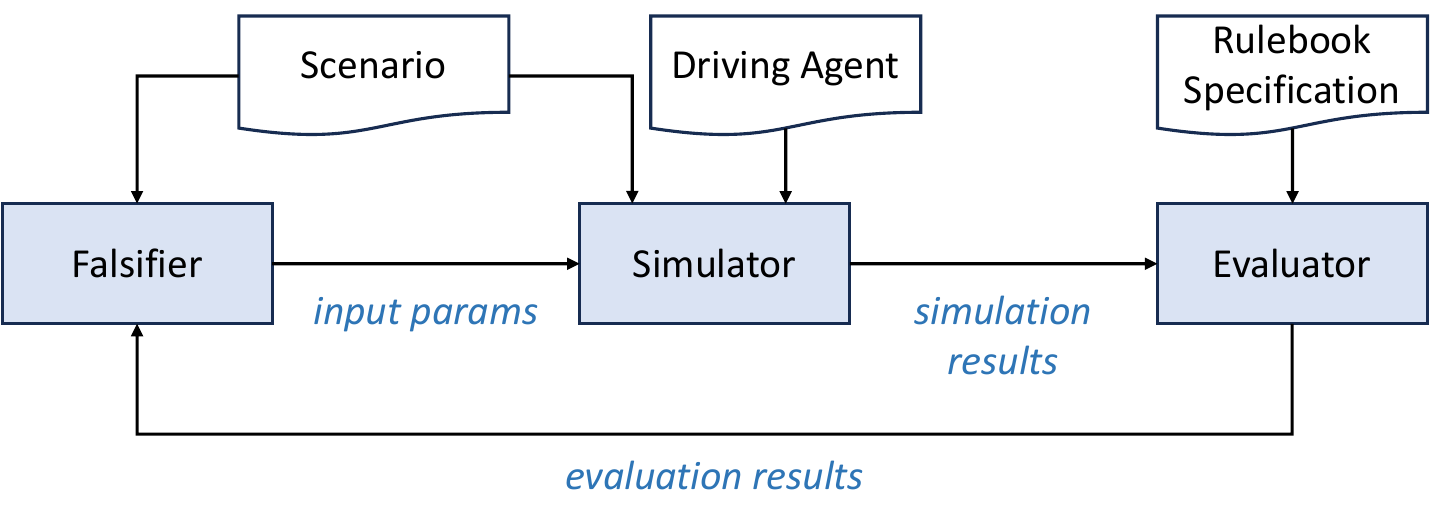}
    \caption{The falsification flow.}
    \label{fig:falsification_flow}
    \vspace{-1mm}
\end{figure}

\begin{table}[t]
    \centering
    \caption{Falsification results on our benchmark.}
    \label{tab:falsification_results}
    \resizebox{\columnwidth}{!}{
    \begin{tabular}{|cc|ccccc|}
    \hline
      \textbf{Scenario} &
      \textbf{Policy} &
      \textbf{\begin{tabular}[c]{@{}c@{}}Avg. \\ EV\end{tabular}} &
      \textbf{\begin{tabular}[c]{@{}c@{}}Max \\ EV\end{tabular}} &
      \textbf{\begin{tabular}[c]{@{}c@{}}CE \\ Ratio\end{tabular}} &
      \textbf{\begin{tabular}[c]{@{}c@{}}Violated \\ Rules (\%)\end{tabular}} &
      \textbf{\begin{tabular}[c]{@{}c@{}}\#Unique\\ Viols.\end{tabular}} \\ \hline\hline
      \multirow{2}{*}{\textbf{\begin{tabular}[c]{@{}c@{}}Common\\ no VRU\end{tabular}}}     & rule-based & $0.127$ & $0.979$ & $76.6\%$ & $92.9\% $ & $168$ \\ 
                                                                                            & ppo        & $0.134$ & $0.995$ & $96.9\%$ & $100.0\%$ & $205$ \\ \hline
      \multirow{2}{*}{\textbf{\begin{tabular}[c]{@{}c@{}}Common\\ with VRU\end{tabular}}}   & rule-based & $0.124$ & $0.979$ & $86.4\%$ & $94.7\% $ & $291$ \\ 
                                                                                            & ppo        & $0.161$ & $0.933$ & $97.7\%$ & $100.0\%$ & $350$ \\ \hline
      \multirow{2}{*}{\textbf{\begin{tabular}[c]{@{}c@{}}Near-accident\\ no VRU\end{tabular}}}  & rule-based & $0.470$ & $0.999$ & $99.3\%$ & $92.9\% $ & $144$ \\ 
                                                                                                & ppo        & $0.252$ & $0.979$ & $96.5\%$ & $100.0\%$ & $100$ \\ \hline
      \multirow{2}{*}{\textbf{\begin{tabular}[c]{@{}c@{}}Near-accident\\ with VRU\end{tabular}}}& rule-based & $0.265$ & $0.658$ & $98.0\%$ & $94.7\% $ & $32 $ \\ 
                                                                                                & ppo        & $0.105$ & $0.657$ & $92.7\%$ & $100.0\%$ & $40 $ \\ \hline
    \end{tabular}
    }
    \begin{tablenotes}
    \item EV: The error value as defined in Definition~\ref{def:error_value}, normalized to $[0, 1]$. The value is 1 if all the rules are violated, and 0 if no rule is violated; CE: A sample is a counterexample if at least one rule is violated; Violated Rules: Percentage of rules violated in at least one sample; \#Unique Viols.: Number of distinct combinations of violated rules.
    \end{tablenotes}
    \vspace{-4mm}
\end{table}

\paragraph{Scenario Coverage and Representativeness}
We first evaluate the representativeness of the selected scenarios. We consider two scenario configurations: an ego vehicle with two other vehicles, and an ego vehicle with two other vehicles plus one pedestrian. Using the Scenic program encoding and the k-center greedy selection method described in Sec.~\ref{subsec:scenarios}, we select 100 representative scenarios for each configuration. Table~\ref{tab:scenario_coverage_results} reports the coverage results.

Taking the second configuration as an example. the Scenic encoding uses 7 dimensions (one for the ego and two for each additional agent). Among the 3232 distinct feasible encodings, the maximum Hamming distance between any of the 3132 unselected encodings and their nearest selected representative is only 2, confirming that every scenario in the space is close to at least one chosen representative. The final four columns further show that all ego and adversary maneuver types, adversary spatial relations, and combinations of adversary maneuvers with spatial relations are fully covered by the selected scenarios. The first configuration yields similar results. Overall, our method produces a compact yet representative set of scenarios that comprehensively spans diverse maneuvers and spatial relations.

\paragraph{Falsification Effectiveness}
Next, we apply simulation-based falsification to assess how effectively the selected scenarios evaluate driving agents under diverse, prioritized objectives. The overall process is summarized in Fig.~\ref{fig:falsification_flow}. As described in Sec.~\ref{subsec:scenic}, semantic parameters in Scenic are defined over continuous ranges. In each iteration, the falsifier samples a parameter assignment;
the scenario is then simulated based on the sampled parameters, using the driving agent under test as the ego.
The resulting trajectory is evaluated against the Rulebook, and the evaluation outcome is fed back to guide subsequent sampling.

We apply falsification to the 200 common scenarios introduced above and the 27 near-accident scenarios constructed from collision reports. Two driving agents are tested: a rule-based planner tracking a given trajectory via PID control, and a Proximal Policy Optimization (PPO)~\cite{schulman2017proximal}-trained policy from the MetaDrive simulator~\cite{li2021metadrive}. All experiments are conducted in MetaDrive. For each scenario, 30 parameter samples are generated using a multi-armed bandit falsifier~\cite{chang2024dynamic, viswanadha2021parallel} interfaced with the VerifAI toolkit~\cite{dreossi2019verifai}. Each iteration takes approximately 20 seconds.

Table~\ref{tab:falsification_results} presents the results. Based on the average error values (EV) and counterexample (CE) ratios, our benchmark successfully identifies rule violations in most scenarios for both agents. The maximum EV values further indicate the benchmark's ability to uncover extreme cases where nearly all rules are violated. We also report the percentage of violated rules—that is, the fraction of rules violated at least once across the sampled executions. All rules are violated in at least one scenario, validating that our benchmark can thoroughly exercise every objective in the rulebook.

Beyond individual rule violations, our benchmark reveals complex joint failure modes. The discovery of hundreds of distinct rule violation combinations indicates that the benchmark exposes not just isolated failures, but also trade-offs among multiple rules. 
Fig.~\ref{fig:failure_modes} illustrates this with two instances of an identical scenario: the ego proceeding straight through an intersection while an oncoming vehicle turns left. In Fig.~\ref{fig:failure_mode_1}, the ego fails to yield and collides, violating collision and clearance rules. In contrast, in Fig.~\ref{fig:failure_mode_2}, the ego swerves to avoid collision but violates drivable area, correct side, and lane-keeping rules.
This contrast demonstrates that our benchmark can uncover qualitatively different failure modes under identical contexts, providing comprehensive coverage of agent shortcomings.

%% file: sections_arxiv/06_conclusion.tex

\section{Conclusion}\label{sec:conclusion}

This work presents \textit{ScenicRules}, the first autonomous driving benchmark that supports multi-objective specifications with explicit priority relations and abstract scenarios. We provide a diverse collection of formalized driving objectives, along with a hierarchical Rulebook structure that encodes priorities in an interpretable and adaptable way. We also construct a compact yet comprehensive set of Scenic scenarios for evaluating driving systems under diverse and challenging conditions. We present experimental evidence that the benchmark is aligned with real-world driving data and can be useful in evaluating autonomous driving systems. 

A primary future direction is to expand the rule set, incorporating both variations of existing objectives and new rule categories. Additional directions include incorporating objectives for multiple interacting agents, scaling \textit{ScenicRules} to more complex scenarios, evaluating advanced agents, and extending our methodology to other types of AI-enabled autonomous systems.

\begin{figure}[t]
    \centering
    \begin{subfigure}[t]{0.9\columnwidth}
        \centering
        \includegraphics[width=\columnwidth]{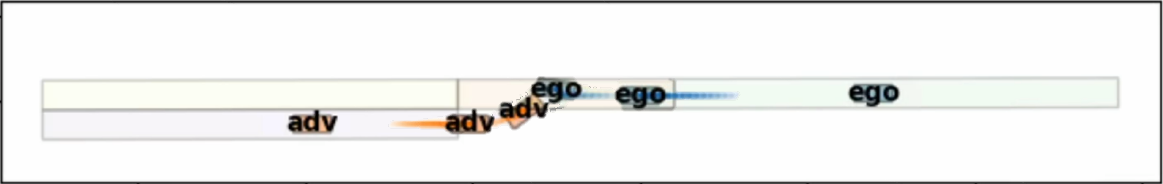}
        \caption{Ego violates \textit{vehicle\_collision}, \textit{front\_clearance}, and \textit{vehicle\_ttc}.}
        \label{fig:failure_mode_1}
        \vspace*{2mm}
    \end{subfigure}
    \begin{subfigure}[t]{0.9\columnwidth}
        \centering
        \includegraphics[width=\columnwidth]{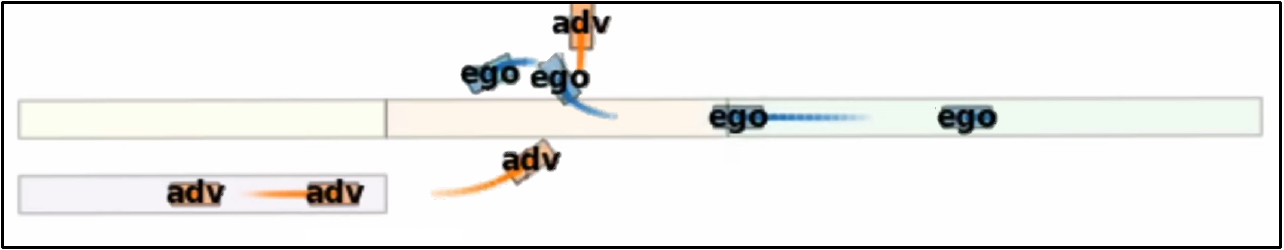}
        \caption{Ego violates \textit{drivable\_area}, \textit{correct\_side}, and \textit{lane\_keeping}.}
        \label{fig:failure_mode_2}
    \end{subfigure}
    \caption{Different failure modes identified under the same driving context (Scenic program).}
    \label{fig:failure_modes}
    \vspace{-4mm}
\end{figure}

%% file: sections_arxiv/appendix.tex
\appendices

\section{Detailed Definitions of Signal Temporal Logic (STL)}\label{appendix:stl}

In this section, we provide the formal definitions of Signal Temporal Logic (STL)~\cite{maler2004monitoring}: Let $\mathbb{T}$ denote the time domain, and $x: \mathbb{T} \rightarrow \mathbb{R}^n$ be a signal. An atomic predicate is a formula of the form $\mu:=f(x(t)) \geq 0$, where $f:\mathbb{R}^n \rightarrow \mathbb{R}$ is a real-valued function.

\begin{definition}[STL Syntax]\label{def:stl_syntax}
The syntax of STL is defined recursively as:
\begin{align}
    \nonumber \varphi := \top \mid \mu \mid \lnot\varphi \mid \varphi_1\land\varphi_2 \mid \varphi_1\cup_I\varphi_2,
\end{align}
where $\varphi, \varphi_1, \varphi_2$ are STL formulas, $\mu$ is an atomic predicate, $\top$ denotes the Boolean constant \textit{true}, and $\cup_I$ is the \textit{until} operator over a nonempty time interval $I$. Based on the definition, the \textit{eventually} operator, $F$, and the \textit{globally} operator, $G$, can be derived as:
\begin{align}
    \nonumber F_{I\varphi} &\equiv \top\cup_I\varphi \\
    \nonumber G_{I\varphi} &\equiv \lnot F_{I} \lnot\varphi
\end{align}
\end{definition}

\begin{definition}[STL Semantics]\label{def:stl_semantics}
The semantics of STL are defined over a signal $x$ at time $t$:
\begin{align}
    \nonumber (x, t) \models \top &\iff \top \\
    \nonumber (x, t) \models \mu &\iff f(x(t)) \geq 0 \\
    \nonumber (x, t) \models \lnot\varphi &\iff (x, t) \not\models \varphi \\
    \nonumber (x, t) \models \varphi_1 \land \varphi_2 &\iff (x, t) \models \varphi_1 \land (x, t) \models \varphi_2 \\ 
    \nonumber (x, t) \models \varphi_1\cup_I\varphi_2 &\iff \exists t'\in t + I \cdot \Big(\big((x, t') \models \varphi_2\big) \\ 
    \nonumber &\phantom{\iff \exists t'} \land \forall t''\in [t, t')\cdot \big((x, t'') \models \varphi_1\big)\Big)
\end{align}
The semantics of the $F$ and $G$ operators can be derived as:
\begin{align}
    \nonumber (x, t) \models F_{I\varphi} &\iff \exists t'\in t + I \cdot \big((x, t') \models \varphi\big) \\
    \nonumber (x, t) \models G_{I\varphi} &\iff \forall t'\in t + I \cdot \big((x, t') \models \varphi\big)
\end{align}
\end{definition}

As mentioned in Sec.~\ref{subsec:stl}, a key advantage of STL is that its formulas can be evaluated quantitatively through a \textit{robustness value}~\cite{donze2010robust, fainekos2009robustness}, denoted as $\rho(\varphi, x, t)$. The value is positive (\textit{resp.} negative) if and only if the formula is satisfied (\textit{resp.} violated). Moreover, larger positive values (\textit{resp.} smaller negative values) indicate stronger satisfaction (\textit{resp.} violation).

\begin{definition}[STL Robustness]\label{def:stl_robustness}
The robustness of an STL formula is defined inductively:
\begin{align}
    \nonumber \rho(\mu, x, t) &\equiv f(x(t)) \\
    \nonumber \rho(\lnot\varphi, x, t) &\equiv -\rho(\varphi, x, t) \\
    \nonumber \rho(\varphi_1 \land \varphi_2, x, t) &\equiv \min(\rho(\varphi_1, x, t), \rho(\varphi_2, x, t)) \\
    \nonumber \rho(\varphi_1\cup_I\varphi_2, x, t) &\equiv \max_{t'\in t+I}\min\big(\rho(\varphi_2, x, t'), \min_{t''\in[t,t')}\rho(\varphi_1, x, t'')\big)
\end{align}

The robustness for $F$ and $G$ can be derived as:
\begin{align}
    \nonumber \rho(F_I\varphi, x, t) &\equiv \max_{t'\in t+I} \rho(\varphi, x, t') \\
    \nonumber \rho(G_I\varphi, x, t) &\equiv \min_{t'\in t+I} \rho(\varphi, x, t')
\end{align}
\end{definition}

\section{Data-driven Priority Optimization}\label{appendix:priority}

As introduced in Sec.~\ref{subsec:our_rulebook}, we propose a data-driven framework to adapt rulebook priorities to different driving contexts. We detail this process in Algorithm~\ref{alg:priority}. Given an initial hierarchical rulebook $\mathcal{B}_M = (\mathcal{R}_M, \preceq_{\mathcal{R}_M})$ and a labeled dataset $L$ containing human preferences, the algorithm outputs an optimized hierarchical rulebook $\mathcal{B}_M^{*}$.

\begin{algorithm}[t]
    \LinesNumbered
    \SetAlgoLined
    \caption{Data-driven Priority Optimization}
    \label{alg:priority}
    
    \KwIn{Initial hierarchical rulebook $\mathcal{B}_M = (\mathcal{R}_M, \preceq_{\mathcal{R}_M})$, labeled dataset $L$}
    \KwOut{Optimized hierarchical rulebook $\mathcal{B}_M^{*}$}
    
    $\mathcal{B}_M^{*} \leftarrow \mathcal{B}_M$\;
    $\kappa^{*} \leftarrow \operatorname{evaluateAgreement}(\mathcal{B}_M^{*}, L)$\;
    $converged \leftarrow \operatorname{False}$\;

    \While{$\neg converged$}{
        $converged \leftarrow \operatorname{True}$\;
        $\mathcal{B}_{best} \leftarrow \mathcal{B}_M^{*}$\;
        $\kappa_{best} \leftarrow \kappa^{*}$\;

        \tcp{Iterate over all rule group pairs}
        \ForEach{$\mathcal{S}_i, \mathcal{S}_j \in \mathcal{R}_M, i < j$}{
            $\tilde{\mathcal{B}_M} \leftarrow \operatorname{swap}(\mathcal{B}_M^{*}, \mathcal{S}_i, \mathcal{S}_j)$\;
            $\tilde{\kappa} \leftarrow \operatorname{evaluateAgreement}(\tilde{\mathcal{B}_M}, L)$\;
            
            \If{$\tilde{\kappa} > \kappa_{best}$}{
                $\mathcal{B}_{best} \leftarrow \tilde{\mathcal{B}_M}$\;
                $\kappa_{best} \leftarrow \tilde{\kappa}$\;
                $converged \leftarrow \operatorname{False}$\;
            }
        }
        $\mathcal{B}_M^{*} \leftarrow \mathcal{B}_{best}$\;
        $\kappa^{*} \leftarrow \kappa_{best}$\;
    }
    \Return{$\mathcal{B}_M^{*}$\;}
\end{algorithm}

The algorithm employs an iterative greedy strategy. First, it calculates the baseline agreement rate between the initial rulebook and the human-labeled dataset (Line 2) using the $\operatorname{evaluateAgreement}$ function. In each iteration of the main loop (Lines 4--19), the algorithm iterates over all unique pairs of rule groups in $\mathcal{R}_M$ (Lines 8--16). For each pair, it tentatively swaps their priority relation using the $\operatorname{swap}$ function (Line 9) and computes the resulting agreement rate (Line 10). If a swap yields a rate higher than the current best (Lines 11--15), the configuration is updated (Lines 17--18). This process repeats until convergence; that is, until a full pass through all pairs yields no further improvement. Finally, the optimized rulebook $\mathcal{B}_M^{*}$ is returned.

We also propose a conservative variant of this process, where only \textit{adjacent} rule groups are swapped (i.e., restricting the loop condition on Line 8 to $j = i + 1$). This constraint preserves the global structure of the initial hierarchy, making it suitable when the initial expert-defined priorities are deemed largely reliable. As shown in our experimental results (Sec.~\ref{subsec:rq1}c), both the standard and conservative variants significantly outperform the initial rulebook. Furthermore, the algorithm converges within two iterations in the experiments, demonstrating its efficiency in practice.

\section{Rule Parameter Optimization}\label{appendix:parameter}

As mentioned in Section~\ref{subsec:rq1}b, parameters within rule formalizations significantly influence evaluation outcomes. To systematically tune these values, we propose a greedy parameter optimization algorithm, detailed in Algorithm~\ref{alg:parameter}. Because higher-priority rules dominate preference comparisons, the algorithm first performs a topological sort of the rules based on their priority relations $\preceq_{\mathcal{R}}$ (Line 2). This ensures that parameters in higher-priority rules are optimized before those in lower-priority ones. For each rule $r$ in topological order (Lines 3--17), if the rule contains tunable parameters $\theta_r$ (Line 4), the algorithm performs a greedy search over candidate parameter values (Lines 7--14). For each candidate value $\tilde{\theta}_r$, it creates a tentative rulebook $\tilde{\mathcal{B}}$ via the $\operatorname{updateParameter}$ function (Line 8) and evaluates the agreement rate with the labeled dataset (Line 9). If this rate exceeds the current maximum (Lines 10--12), the optimal parameter configuration is updated. After evaluating all candidates for rule $r$, the rulebook is updated with the optimal parameter value $\theta_r^{*}$ (Line 15). This sequential process continues until all rules are processed, ultimately returning the fully optimized rulebook $\mathcal{B}^{*}$ (Line 18). Note that this approach assumes a finite set of candidate values for each parameter.

\begin{algorithm}[t]
    \LinesNumbered
    \SetAlgoLined
    \caption{Rule Parameter Optimization}
    \label{alg:parameter}
    
    \KwIn{Rulebook $\mathcal{B} = (\mathcal{R}, \preceq_{\mathcal{R}})$, labeled dataset $L$}
    \KwOut{Parameter-optimized rulebook $\mathcal{B}^{*}$}
    
    $\mathcal{B}^{*} \leftarrow \mathcal{B}$\;
    $\mathcal{T} \leftarrow \operatorname{topologicalSort}(\mathcal{R}, \preceq_{\mathcal{R}})$\;
    
    \tcp{Traverse rule hierarchy in topological order}
    \ForEach{$r \in \mathcal{T}$}{
        \If{$r$ has parameters $\theta_r$}{
            $\kappa^{*} \leftarrow -\infty$\;
            $\theta_r^{*} \leftarrow \theta_r$\;
            
            \tcp{Greedy search over parameter space}
            \ForEach{$\tilde{\theta}_r \in \operatorname{candidateValues}(\theta_r)$}{
                $\tilde{\mathcal{B}} \leftarrow \operatorname{updateParameter}(\mathcal{B}^{*}, r, \tilde{\theta}_r)$\;
                $\tilde{\kappa} \leftarrow \operatorname{evaluateAgreement}(\tilde{\mathcal{B}}, L)$\;
                
                \If{$\tilde{\kappa} > \kappa^{*}$}{
                    $\kappa^{*} \leftarrow \tilde{\kappa}$\;
                    $\theta_r^{*} \leftarrow \tilde{\theta}_r$\;
                }
            }

            $\mathcal{B}^{*} \leftarrow \operatorname{updateParameter}(\mathcal{B}^{*}, r, \theta_r^{*})$\;
        }
    }
    \Return{$\mathcal{B}^{*}$\;}
\end{algorithm}

\section{Common Scenario Generation}\label{appendix:common_scenario}

As described in Sec.~\ref{subsubsec:common_scenarios}, we develop an automated Scenic program generator (Fig.~\ref{fig:scenario_generator}) that produces Scenic programs based on specified agent maneuvers and spatial relations. To illustrate the generator's mechanics, we provide an example input specification and its corresponding output.

Fig.~\ref{fig:scenario_spec} presents a sample common scenario specification in a JSON-style format. It defines the scenario name (Line 2), the map (Line 3), the ego vehicle's maneuver (Line 4), and the maneuvers and spatial relations for surrounding agents (Lines 5--21). The resulting Scenic program is displayed in Figs.~\ref{fig:scenic_code_common_1} and~\ref{fig:scenic_code_common_2}. Leveraging the structured nature of Scenic, our generator systematically constructs the program by defining the following components:
\begin{itemize}
    \item \textsc{Map and Model}: Specifies the map file path and the Scenic domain.
    \item \textsc{Parameters and Constants}: Defines semantic parameters (including their valid ranges) and global constants.
    \item \textsc{Spatial Relations}: Sets the initial positions of agents and the spatial relations between them.
    \item \textsc{Agent Behaviors}: Defines the dynamic behaviors and control policies of the agents.
    \item \textsc{Specifications}: Create agent instances and links them to their assigned initial positions and behaviors.
\end{itemize}

\section{LLM-assisted Scenic Program Generation}\label{appendix:llm}

As described in Sec.~\ref{subsubsec:near_accident_scenarios}, we leverage large language models (LLMs) to convert one-paragraph natural language descriptions into corresponding Scenic programs. Figs.~\ref{fig:prompt_1} and~\ref{fig:prompt_2} detail the prompts used for the LLM. The prompt structure begins with an overview of Scenic and its architecture. Next, we outline the key syntax elements relevant to each section of a Scenic program. We then provide several few-shot examples consisting of $\langle$natural language description, Scenic program$\rangle$ pairs. Finally, we append the natural language description of the target collision report.

While most generated programs closely adhere to the Scenic structure and syntax, minor errors persist, particularly involving non-trivial spatial relations or complex agent behaviors. A key direction for future work is the refinement of this LLM-based code generation pipeline.

\section{Evaluation Flow Against Human Annotations}\label{appendix:eval_human}

\begin{figure}[t]
    \centering
    \includegraphics[width=\linewidth]{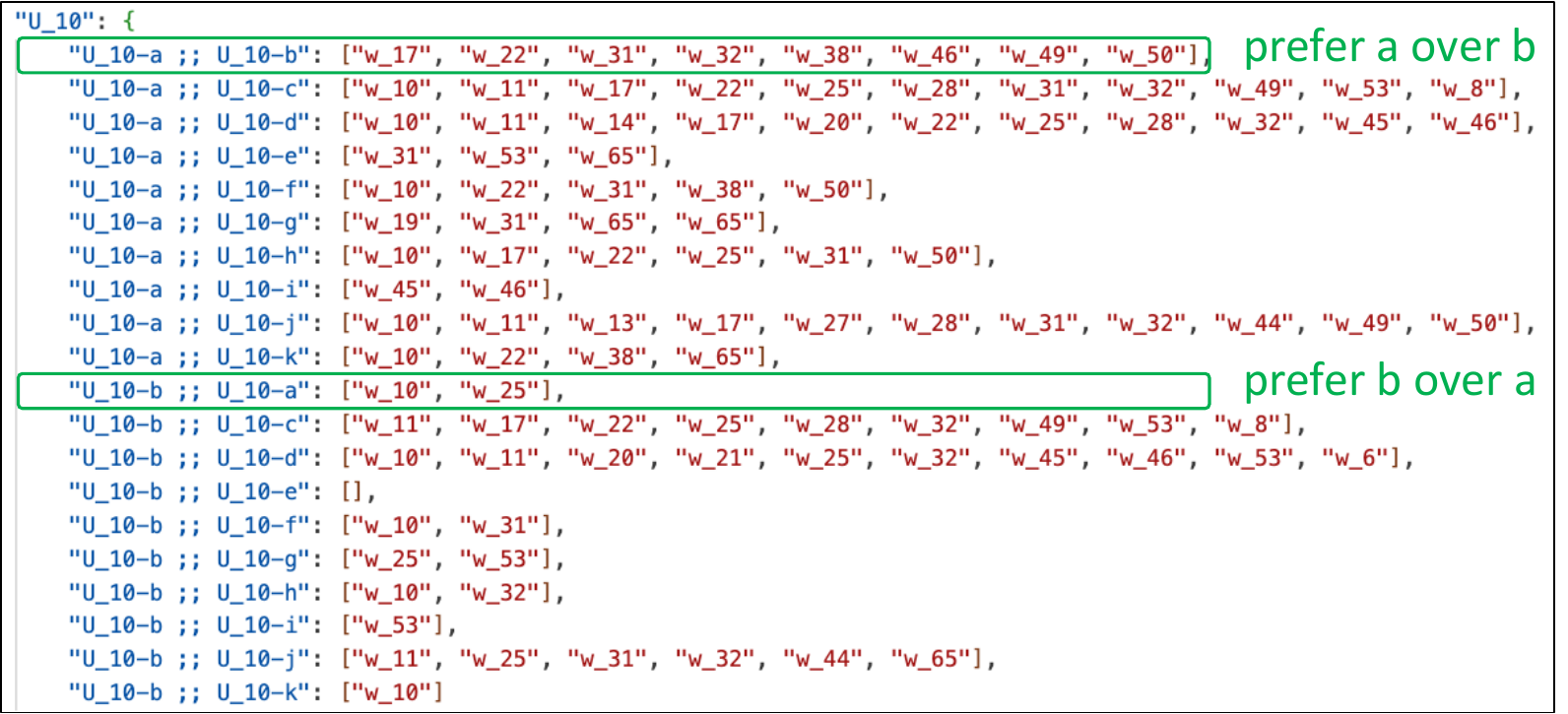}
    \caption{Example from the Reasonable Crowd dataset~\cite{helou2021reasonable}. Each scenario contains multiple feasible trajectories, and human annotators provide pairwise preference labels.}
    \label{fig:reasonable_crowd}
\end{figure}

In our experimental results (Sec.~\ref{subsec:rq1}), we evaluate the alignment between rulebook-induced preferences and human-derived preferences. This section details the experimental methodology. Fig.~\ref{fig:reasonable_crowd} displays a snippet from the Reasonable Crowd dataset~\cite{helou2021reasonable}. In this example, ``$U_{10}$'' represents a scenario comprising several feasible trajectories ($U_{10-a}$ through $U_{10-k}$). Each entry corresponds to an ordered pair of trajectories, recording the count of human annotators who preferred the first trajectory over the second. For instance, eight annotators favored trajectory $U_{10-a}$ over $U_{10-b}$, while two preferred the reverse. 

\begin{figure}[t]
    \centering
    \includegraphics[width=\linewidth]{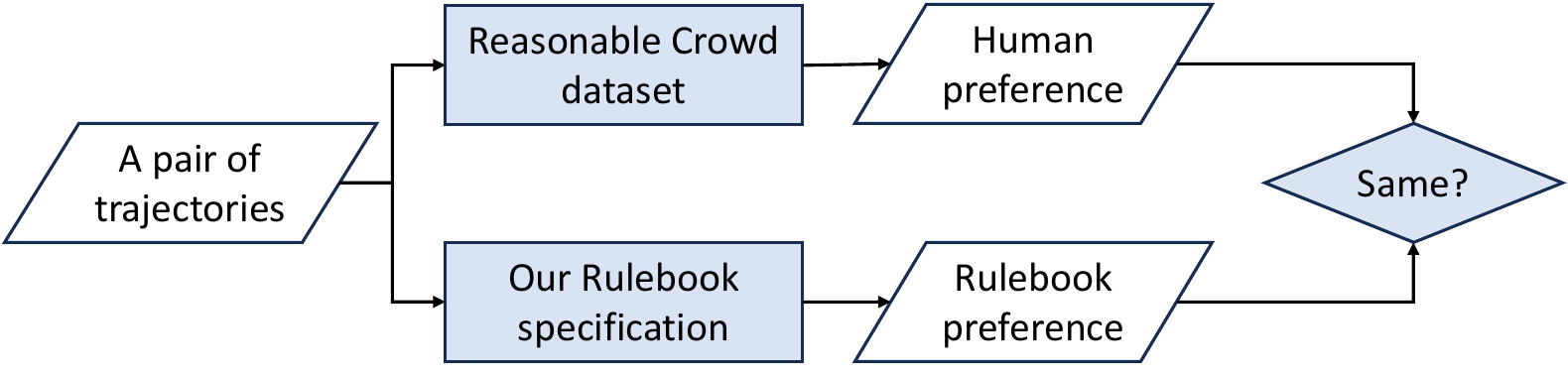}
    \caption{Evaluation flow for comparing hierarchical Rulebook preferences with human preferences.}
    \label{fig:reasonable_crowd_eval_flow}
\end{figure}

Fig.~\ref{fig:reasonable_crowd_eval_flow} illustrates the complete evaluation pipeline. Adopting the methodology from~\cite{helou2021reasonable}, we employ the Bradley–Terry model~\cite{bradley1952rank} to estimate the ground-truth human preference for each trajectory pair in the dataset. Simultaneously, our rulebook determines its preference by analyzing the rule violations of both trajectories in the context of the defined priority relations. Finally, we compare the rulebook's preference against the human consensus to compute the overall accuracy.

\begin{figure*}[t]
\centering
\begin{lstlisting}[language=json]
{
    "scenario": "common.scenic",
    "map": "Town05.xodr",
    "ego": {"type": "CAR", "maneuver": "LANE_FOLLOWING"},
    "agents": {
        "car1": {
            "type": "CAR", 
            "maneuver": "RIGHT_TURN",
            "spatial_relation": "OPPOSITE_INTERSECTION"
        },
        "car2": {
            "type": "CAR", 
            "maneuver": "LANE_FOLLOWING",
            "spatial_relation": "LATERAL_INTERSECTION"
        },
        "ped1": {
            "type": "PEDESTRIAN", 
            "maneuver": "CROSS_STREET",
            "spatial_relation": "SIDEWALK"
        }
    }
}
\end{lstlisting}
\caption{An example common scenario specification.}
\label{fig:scenario_spec}
\end{figure*}

\begin{figure*}[t]
\centering
\begin{lstlisting}[style=pythonstyle]
#################################
# MAP AND MODEL                 #
#################################
param map = localPath('maps/Town05.xodr')
model scenic.domains.driving.model
param POLICY = 'built_in'

#################################
# PARAMETERS AND CONSTANTS      #
#################################
param PED1_LONGITUDINAL_OFFSET = VerifaiRange(5, 15)
param EGO_SPEED = VerifaiRange(6, 11)
param EGO_BRAKE = VerifaiRange(0.5, 1.0)
param EGO_SAFETY_DIST = VerifaiRange(6, 10)
param CAR1_SPEED = VerifaiRange(6, 11)
param CAR2_SPEED = VerifaiRange(6, 11)
param PED1_SPEED = VerifaiRange(1, 3)

MODEL = 'vehicle.lincoln.mkz_2017'
CAR1_INIT_DIST = [15, 25]
CAR2_INIT_DIST = [15, 25]
PED1_LATERAL_OFFSET = 8

#################################
# SPATIAL RELATIONS             #
#################################
intersection = Uniform(*filter(lambda i: i.is4Way, network.intersections))

egoInitLane = Uniform(*intersection.incomingLanes)
egoSpawnPt = new OrientedPoint in egoInitLane.centerline

car1InitLane = Uniform(*filter(lambda m:
    m.type is ManeuverType.STRAIGHT,
    Uniform(*filter(lambda m: 
        m.type is ManeuverType.STRAIGHT, 
        egoInitLane.maneuvers)
    ).reverseManeuvers)
).startLane
car1SpawnPt = new OrientedPoint in car1InitLane.centerline

car2InitLane = Uniform(*filter(lambda m:
    m.type is ManeuverType.STRAIGHT,
    Uniform(*filter(lambda m: 
        m.type is ManeuverType.STRAIGHT, 
        egoInitLane.maneuvers)
    ).conflictingManeuvers)
).startLane
car2SpawnPt = new OrientedPoint in car2InitLane.centerline

ped1SpawnPt = new OrientedPoint at egoSpawnPt offset by (PED1_LATERAL_OFFSET, globalParameters.PED1_LONGITUDINAL_OFFSET, 0)
ped1EndPt = new OrientedPoint at egoSpawnPt offset by (-PED1_LATERAL_OFFSET, globalParameters.PED1_LONGITUDINAL_OFFSET, 0)
\end{lstlisting}
\caption{The corresponding Scenic program of the specification in Fig.~\ref{fig:scenario_spec} (Part I).}
\label{fig:scenic_code_common_1}
\end{figure*}

\begin{figure*}[t]
\centering
\begin{lstlisting}[style=pythonstyle]
#################################
# AGENT BEHAVIORS               #
#################################
behavior EgoBehavior():
    try:
        do FollowLaneBehavior(target_speed=globalParameters.EGO_SPEED)
    interrupt when withinDistanceToObjsInLaneNoInter(self, globalParameters.EGO_SAFETY_DIST):
        take SetBrakeAction(globalParameters.EGO_BRAKE)

car1Maneuver = Uniform(*filter(lambda m: m.type is ManeuverType.RIGHT_TURN, car1InitLane.maneuvers))
car1Trajectory = [car1InitLane, car1Maneuver.connectingLane, car1Maneuver.endLane]
behavior Car1Behavior(trajectory):
    do FollowTrajectoryBehavior(target_speed=globalParameters.CAR1_SPEED, trajectory=trajectory)

behavior Car2Behavior():
    do FollowLaneBehavior(target_speed=globalParameters.CAR2_SPEED)

behavior Ped1Behavior():
    take SetWalkingSpeedAction(speed=globalParameters.PED1_SPEED)

#################################
# SPECIFICATIONS                #
#################################
ego  = new Car at egoSpawnPt,
      with blueprint MODEL,
      with behavior EgoBehavior()

car1 = new Car at car1SpawnPt,
      with blueprint MODEL,
      with behavior Car1Behavior(car1Trajectory)

car2 = new Car at car2SpawnPt,
      with blueprint MODEL,
      with behavior Car2Behavior()

ped1 = new Pedestrian at ped1SpawnPt, facing toward ped1EndPt,
      with behavior Ped1Behavior()
\end{lstlisting}
\caption{The corresponding Scenic program of the specification in Fig.~\ref{fig:scenario_spec}  (Part II).}
\label{fig:scenic_code_common_2}
\end{figure*}

\begin{figure*}[t]
\centering
\begin{lstlisting}[style=promptstyle]
You are an expert in the Scenic programming language. Your task is to convert natural language descriptions of scenarios into valid Scenic code.
A Scenic program consists of several sections, including:
- 'MAP AND MODEL': for specifying the map and driving models used in the scenario
- 'CONSTANTS': for defining parameters and constants
- 'AGENT BEHAVIORS': for defining bahaviors of agents in the scenario
- 'SPATIAL RELATIONS': for specifying the initial positions and trajectories (if needed) of agents
- 'SCENARIO SPECIFICATION': for defining the object instances with their behaviors
Each section is clearly marked with a header (e.g., '#################################').
When writing the Scenic code, ensure that the code is syntactically correct and all necessary sections are included and properly formatted.
Below is a detailed syntax guide for every section. PLEASE STRICTLY FOLLOW THE SYNTAX. DO NOT HALLUCINATE.
### MAP AND MODEL ###
In this section, just paste the following three lines:
param map = localPath('../../maps/Town04.xodr')
model scenic.simulators.metadrive.model
param POLICY = 'built_in'
### CONSTANTS ###
In this section, define all the constants and parameters used in the scenario.
Constants can be defined using simple assignments (e.g., MODEL = 'vehicle.lincoln.mkz_2017', EGO_INIT_DIST = [20, 25]).
Parameters that need to be sampled during scenario generation can be defined using the 'param' keyword along with 'VerifaiRange' for continuous ranges (e.g., param EGO_SPEED = VerifaiRange(7, 10)). Later when using these parameters in the scenario, refer to them as 'globalParameters.PARAM_NAME'.
### AGENT BEHAVIORS ###
In this section, define the behaviors of all the agents involved in the scenario.
Each behavior should be defined using the 'behavior' keyword followed by the behavior name and parameters (if any).
You can adopt the built-in behaviors and actions provided by Scenic.
When using a built-in behavior, use the 'do' keyword followed by the behavior name and its parameters. These are the built-in behaviors you can use:
- FollowTrajectoryBehavior(target_speed, trajectory)
- FollowLaneBehavior(target_speed, laneToFollow=None)
- LaneChangeBehavior(laneSectionToSwitch, target_speed)
When using a built-in action, use the 'take' keyword followed by the action name and its parameters. These are the built-in actions you can use:
- SetBrakeAction(brake)
- SetSteerAction(steer)
- SetThrottleAction(throttle)
- SetSpeedAction(speed)
- SetVelocityAction(xVel, yVel)
You can also use try-interrupt statements to define complex behaviors. The 'try' block contains the main behavior, while the 'interrupt when' clauses specify conditions that can interrupt the main behavior and trigger alternative actions or behaviors.
Here is an example of defining a behavior with try-interrupt statements:
behavior EgoBehavior(trajectory):
    try:
        do FollowTrajectoryBehavior(target_speed=globalParameters.EGO_SPEED, trajectory=trajectory)
    interrupt when withinDistanceToAnyObjs(self, globalParameters.SAFETY_DIST):
        take SetBrakeAction(globalParameters.EGO_BRAKE)
    interrupt when withinDistanceToAnyObjs(self, CRASH_DIST):
        terminate
Here are some predefined functions or operators you may use in the interrupt conditions:
- withinDistanceToAnyObjs(vehicle, distance): checks if the agent is within a certain distance to any other objects.
- withinDistanceToObjsInLane(vehicle, distance): checks whether there exists any obj (1) in front of the vehicle, (2) on the same lane, and (3) within the threshold distance.
- distance [from vector] to vector: computes the distance to the given position from ego (or the position provided with the optional from vector)
- angle [from vector] to vector: computes the heading (azimuth) to the given position from ego (or the position provided with the optional from vector). For example, if angle to taxi is zero, then taxi is due North of ego.
- apparent heading of vector [from vector]: computes the apparent heading of the vector, with respect to the line of sight from ego (or the position provided with the optional from vector)
- vector in region: checks if the given position is inside the specified region
\end{lstlisting}
\caption{The prompt used for LLM-assisted Scenic program generation (Part I).}
\label{fig:prompt_1}
\end{figure*}

\begin{figure*}[t]
\centering
\begin{lstlisting}[style=promptstyle]
### SPATIAL RELATIONS ###
In this section, define the initial positions and trajectories (if needed) of the agents.
First, the hierarchy of road network structure is road -> laneGroup -> lane -> laneSection. A 'road' consists of multiple 'laneGroups', each 'laneGroup' contains multiple 'lanes', and each 'lane' is divided into multiple 'laneSections'.
At each level, you can access its ancestors and descendants (descendants may be a list).
In addition, for lanes, you can access its maneuvers, which represent possible driving actions (e.g., straight, left turn, right turn) that can be taken from the end of the lane.
For laneSections (note not for lanes), you can access its fasterLane and slowerLane (if any) to facilitate lane change maneuvers.
When defining initial positions, you can use constructs like 'new OrientedPoint in lane.centerline' to specify a random point along the centerline of a lane.
You can also use relative positioning to define positions based on other agents or points, such as 'new OrientedPoint following roadDirection from egoSpawnPt for ADV_DIST_FROM_EGO_INIT'.
Here are the available specifiers you can use:
- offset by vector: Positions the object at the given coordinates in the local coordinate system of ego (which must already be defined).
- offset along direction by vector: Positions the object at the given coordinates, in a local coordinate system centered at ego and oriented along the given direction.
- following vectorField [from vector] for scalar: Position by following the given vector field for the given distance starting from ego or the given vector.
- (ahead of | behind) vector [by scalar]: Positions the object to the front/back by the given scalar distance
- (left of | right of) vector [by scalar]: Positions the object to the left/right by the given scalar distance
When defining trajectories, you can create a list of lanes that the agent will follow during the scenario.
### SCENARIO SPECIFICATION ###
In this section, define the object instances involved in the scenario along with their behaviors.
Each object instance should be created using the 'new' keyword followed by the object type (e.g., Car) and its initial position.
You can specify additional properties for each object, such as the blueprint model and the behavior to be used.
Here is an example of defining an ego vehicle and an adversary vehicle:
ego = new Car at egoSpawnPt,
    with blueprint MODEL,
    with behavior EgoBehavior(egoTrajectory)
adversary = new Car at advSpawnPt,
    with blueprint MODEL,
    with behavior FollowLaneBehavior(target_speed=globalParameters.ADV_SPEED)
You can also include requirements and termination conditions for the scenario using 'require' and 'terminate when' statements.
Note that within requirements and termination conditions parameters cannot be accessed, only constants can be used.

Here are three examples of natural language descriptions with their corresponding Scenic codes:
Natural Language Description 1: 
[DESCRIPTION]
Scenic Code 1:
[SCENIC CODE]

Natural Language Description 2: 
[DESCRIPTION]
Scenic Code 2:
[SCENIC CODE]

Natural Language Description 3: 
[DESCRIPTION]
Scenic Code 3:
[SCENIC CODE]

Now, convert the following natural language description into a Scenic program:
[DESCRIPTION]

Please just return the Scenic program following the format of the provided example. DO NOT include any additional text or explanations. DO NOT ADD COMMENTS in the generated Scenic program!
\end{lstlisting}
\caption{The prompt used for LLM-assisted Scenic program generation (Part II).}
\label{fig:prompt_2}
\end{figure*}

%% file: arxiv.bbl
\begin{thebibliography}{10}
\providecommand{\url}[1]{#1}
\csname url@samestyle\endcsname
\providecommand{\newblock}{\relax}
\providecommand{\bibinfo}[2]{#2}
\providecommand{\BIBentrySTDinterwordspacing}{\spaceskip=0pt\relax}
\providecommand{\BIBentryALTinterwordstretchfactor}{4}
\providecommand{\BIBentryALTinterwordspacing}{\spaceskip=\fontdimen2\font plus
\BIBentryALTinterwordstretchfactor\fontdimen3\font minus \fontdimen4\font\relax}
\providecommand{\BIBforeignlanguage}[2]{{%
\expandafter\ifx\csname l@#1\endcsname\relax
\typeout{** WARNING: IEEEtran.bst: No hyphenation pattern has been}%
\typeout{** loaded for the language `#1'. Using the pattern for}%
\typeout{** the default language instead.}%
\else
\language=\csname l@#1\endcsname
\fi
#2}}
\providecommand{\BIBdecl}{\relax}
\BIBdecl

\bibitem{kxan2024waymo}
\BIBentryALTinterwordspacing
{KXAN Austin}, ``How {Waymo's} driverless technology avoided scooter rider who fell into {Austin} road,'' 2024. [Online]. Available: \url{https://youtu.be/h7PGrAlPELc?si=ZkURWsldNjSi3OKQ}
\BIBentrySTDinterwordspacing

\bibitem{elluswamy2025tesla}
\BIBentryALTinterwordspacing
A.~Elluswamy, ``Tesla's approach to autonomy,'' 2025. [Online]. Available: \url{https://x.com/aelluswamy/status/1981644831790379245}
\BIBentrySTDinterwordspacing

\bibitem{helou2021reasonable}
B.~Helou, A.~Dusi, A.~Collin, N.~Mehdipour, Z.~Chen, C.~Lizarazo, C.~Belta, T.~Wongpiromsarn, R.~D. Tebbens, and O.~Beijbom, ``The {Reasonable Crowd}: Towards evidence-based and interpretable models of driving behavior,'' in \emph{IEEE/RSJ International Conference on Intelligent Robots and Systems (IROS)}, 2021, pp. 6708--6715.

\bibitem{chang2024dynamic}
K.~K.-C. Chang, K.~Xu, E.~Kim, A.~Sangiovanni-Vincentelli, and S.~A. Seshia, ``Dynamic, multi-objective specification and falsification of autonomous {CPS},'' in \emph{International Conference on Runtime Verification (RV)}.\hskip 1em plus 0.5em minus 0.4em\relax Springer, 2024, pp. 40--58.

\bibitem{viswanadha2021addressing}
K.~Viswanadha, F.~Indaheng, J.~Wong, E.~Kim, E.~Kalvan, Y.~Pant, D.~J. Fremont, and S.~A. Seshia, ``Addressing the {IEEE AV} test challenge with {Scenic} and {VerifAI},'' in \emph{IEEE International Conference on Artificial Intelligence Testing (AITest)}.\hskip 1em plus 0.5em minus 0.4em\relax IEEE, 2021, pp. 136--142.

\bibitem{shalev2017formal}
S.~Shalev-Shwartz, S.~Shammah, and A.~Shashua, ``On a formal model of safe and scalable self-driving cars,'' \emph{arXiv preprint arXiv:1708.06374}, 2017.

\bibitem{bellem2016objective}
H.~Bellem, T.~Sch{\"o}nenberg, J.~F. Krems, and M.~Schrauf, ``Objective metrics of comfort: Developing a driving style for highly automated vehicles,'' \emph{Transportation Research Part F: Traffic Psychology and Behaviour}, vol.~41, pp. 45--54, 2016.

\bibitem{maierhofer2022formalization}
S.~Maierhofer, P.~Moosbrugger, and M.~Althoff, ``Formalization of intersection traffic rules in temporal logic,'' in \emph{IEEE Intelligent Vehicles Symposium (IV)}.\hskip 1em plus 0.5em minus 0.4em\relax IEEE, 2022, pp. 1135--1144.

\bibitem{ettinger2021large}
S.~Ettinger, S.~Cheng, B.~Caine, C.~Liu, H.~Zhao, S.~Pradhan, Y.~Chai, B.~Sapp, C.~R. Qi, Y.~Zhou \emph{et~al.}, ``Large scale interactive motion forecasting for autonomous driving: The {Waymo Open Motion Dataset},'' in \emph{IEEE/CVF International Conference on Computer Vision (ICCV)}, 2021, pp. 9710--9719.

\bibitem{wilson2023argoverse}
B.~Wilson, W.~Qi, T.~Agarwal, J.~Lambert, J.~Singh, S.~Khandelwal, B.~Pan, R.~Kumar, A.~Hartnett, J.~K. Pontes \emph{et~al.}, ``Argoverse 2: Next generation datasets for self-driving perception and forecasting,'' \emph{arXiv preprint arXiv:2301.00493}, 2023.

\bibitem{caesar2020nuscenes}
H.~Caesar, V.~Bankiti, A.~H. Lang, S.~Vora, V.~E. Liong, Q.~Xu, A.~Krishnan, Y.~Pan, G.~Baldan, and O.~Beijbom, ``{nuScenes}: A multimodal dataset for autonomous driving,'' in \emph{IEEE/CVF Conference on Computer Vision and Pattern Recognition (CVPR)}, 2020, pp. 11\,621--11\,631.

\bibitem{houston2021one}
J.~Houston, G.~Zuidhof, L.~Bergamini, Y.~Ye, L.~Chen, A.~Jain, S.~Omari, V.~Iglovikov, and P.~Ondruska, ``One thousand and one hours: Self-driving motion prediction dataset,'' in \emph{Conference on Robot Learning (CoRL)}.\hskip 1em plus 0.5em minus 0.4em\relax PMLR, 2021, pp. 409--418.

\bibitem{karnchanachari2024towards}
N.~Karnchanachari, D.~Geromichalos, K.~S. Tan, N.~Li, C.~Eriksen, S.~Yaghoubi, N.~Mehdipour, G.~Bernasconi, W.~K. Fong, Y.~Guo \emph{et~al.}, ``Towards learning-based planning: The {nuPlan} benchmark for real-world autonomous driving,'' in \emph{IEEE International Conference on Robotics and Automation (ICRA)}.\hskip 1em plus 0.5em minus 0.4em\relax IEEE, 2024, pp. 629--636.

\bibitem{caesar2021nuplan}
H.~Caesar, J.~Kabzan, K.~S. Tan, W.~K. Fong, E.~Wolff, A.~Lang, L.~Fletcher, O.~Beijbom, and S.~Omari, ``{nuPlan}: A closed-loop {ML}-based planning benchmark for autonomous vehicles,'' \emph{arXiv preprint arXiv:2106.11810}, 2021.

\bibitem{althoff2017commonroad}
M.~Althoff, M.~Koschi, and S.~Manzinger, ``{CommonRoad}: Composable benchmarks for motion planning on roads,'' in \emph{IEEE Intelligent Vehicles Symposium (IV)}.\hskip 1em plus 0.5em minus 0.4em\relax IEEE, 2017, pp. 719--726.

\bibitem{dauner2024navsim}
D.~Dauner, M.~Hallgarten, T.~Li, X.~Weng, Z.~Huang, Z.~Yang, H.~Li, I.~Gilitschenski, B.~Ivanovic, M.~Pavone \emph{et~al.}, ``{NAVSIM}: Data-driven non-reactive autonomous vehicle simulation and benchmarking,'' \emph{Advances in Neural Information Processing Systems}, vol.~37, pp. 28\,706--28\,719, 2024.

\bibitem{zhu2025m3cad}
M.~Zhu, Y.~Zhu, Y.~Zhu, Q.~Chen, D.~Qu, S.~Fu, and Q.~Yang, ``{M$^3$CAD}: Towards generic cooperative autonomous driving benchmark,'' \emph{arXiv preprint arXiv:2505.06746}, 2025.

\bibitem{censi2019liability}
A.~Censi, K.~Slutsky, T.~Wongpiromsarn, D.~Yershov, S.~Pendleton, J.~Fu, and E.~Frazzoli, ``Liability, ethics, and culture-aware behavior specification using rulebooks,'' in \emph{International Conference on Robotics and Automation (ICRA)}.\hskip 1em plus 0.5em minus 0.4em\relax IEEE, 2019, pp. 8536--8542.

\bibitem{zhan2019interaction}
W.~Zhan, L.~Sun, D.~Wang, H.~Shi, A.~Clausse, M.~Naumann, J.~Kummerle, H.~Konigshof, C.~Stiller, A.~de~La~Fortelle \emph{et~al.}, ``{INTERACTION} dataset: An international, adversarial and cooperative motion dataset in interactive driving scenarios with semantic maps,'' \emph{arXiv preprint arXiv:1910.03088}, 2019.

\bibitem{prabu2022scendd}
A.~Prabu, N.~Ranjan, L.~Li, R.~Tian, S.~Chien, Y.~Chen, and R.~Sherony, ``{SceNDD}: A scenario-based naturalistic driving dataset,'' in \emph{IEEE 25th International Conference on Intelligent Transportation Systems (ITSC)}.\hskip 1em plus 0.5em minus 0.4em\relax IEEE, 2022, p. 4363–4368.

\bibitem{schuldes2024scenario}
M.~Schuldes, C.~Glasmacher, and L.~Eckstein, ``scenario.center: Methods from real-world data to a scenario database,'' in \emph{IEEE Intelligent Vehicles Symposium (IV)}.\hskip 1em plus 0.5em minus 0.4em\relax IEEE, 2024, pp. 1119--1126.

\bibitem{moser2025coreset}
B.~B. Moser, A.~S. Shanbhag, S.~Frolov, F.~Raue, J.~Folz, and A.~Dengel, ``A coreset selection of coreset selection literature: Introduction and recent advances,'' \emph{arXiv preprint arXiv:2505.17799}, 2025.

\bibitem{sener2017active}
O.~Sener and S.~Savarese, ``Active learning for convolutional neural networks: A core-set approach,'' \emph{arXiv preprint arXiv:1708.00489}, 2017.

\bibitem{fremont2023scenic}
D.~J. Fremont, E.~Kim, T.~Dreossi, S.~Ghosh, X.~Yue, A.~L. Sangiovanni-Vincentelli, and S.~A. Seshia, ``Scenic: A language for scenario specification and data generation,'' \emph{Machine Learning}, vol. 112, no.~10, pp. 3805--3849, 2023.

\bibitem{fremont2019scenic}
D.~J. Fremont, T.~Dreossi, S.~Ghosh, X.~Yue, A.~L. Sangiovanni-Vincentelli, and S.~A. Seshia, ``Scenic: A language for scenario specification and scene generation,'' in \emph{ACM SIGPLAN Conference on Programming Language Design and Implementation (PLDI)}, June 2019.

\bibitem{liu2024survey}
M.~Liu, E.~Yurtsever, J.~Fossaert, X.~Zhou, W.~Zimmer, Y.~Cui, B.~L. Zagar, and A.~C. Knoll, ``A survey on autonomous driving datasets: Statistics, annotation quality, and a future outlook,'' \emph{IEEE Transactions on Intelligent Vehicles (T-IV)}, 2024.

\bibitem{carlachallenge}
\BIBentryALTinterwordspacing
CARLA, ``Carla autonomous driving challenge.'' [Online]. Available: \url{https://leaderboard.carla.org/challenge/}
\BIBentrySTDinterwordspacing

\bibitem{cao2025pseudo}
W.~Cao, M.~Hallgarten, T.~Li, D.~Dauner, X.~Gu, C.~Wang, Y.~Miron, M.~Aiello, H.~Li, I.~Gilitschenski, B.~Ivanovic, M.~Pavone, A.~Geiger, and K.~Chitta, ``Pseudo-simulation for autonomous driving,'' in \emph{Conference on Robot Learning (CoRL)}, 2025.

\bibitem{10051644}
\BIBentryALTinterwordspacing
``Example applications of {IEEE} std 2846-2022 to formal safety-related models,'' 2023. [Online]. Available: \url{https://ieeexplore.ieee.org/document/10051644}
\BIBentrySTDinterwordspacing

\bibitem{collin2020safety}
A.~Collin, A.~Bilka, S.~Pendleton, and R.~D. Tebbens, ``Safety of the intended driving behavior using rulebooks,'' in \emph{IEEE Intelligent Vehicles Symposium (IV)}, 2020, pp. 136--143.

\bibitem{viswanadha2021parallel}
K.~Viswanadha, E.~Kim, F.~Indaheng, D.~J. Fremont, and S.~A. Seshia, ``Parallel and multi-objective falsification with {Scenic} and {VerifAI},'' in \emph{International Conference on Runtime Verification (RV)}.\hskip 1em plus 0.5em minus 0.4em\relax Springer, 2021, pp. 265--276.

\bibitem{seshia-cacm22a}
S.~A. Seshia, D.~Sadigh, and S.~S. Sastry, ``Toward verified artificial intelligence,'' \emph{Communications of the {ACM}}, vol.~65, no.~7, pp. 46--55, 2022.

\bibitem{maler2004monitoring}
O.~Maler and D.~Nickovic, ``Monitoring temporal properties of continuous signals,'' in \emph{International Symposium on Formal Techniques in Real-Time and Fault-Tolerant Systems (FTRTFT)}.\hskip 1em plus 0.5em minus 0.4em\relax Springer, 2004, pp. 152--166.

\bibitem{donze2010robust}
A.~Donz{\'e} and O.~Maler, ``Robust satisfaction of temporal logic over real-valued signals,'' in \emph{International Conference on Formal Modeling and Analysis of Timed Systems (FORMATS)}.\hskip 1em plus 0.5em minus 0.4em\relax Springer, 2010, pp. 92--106.

\bibitem{fainekos2009robustness}
G.~E. Fainekos and G.~J. Pappas, ``Robustness of temporal logic specifications for continuous-time signals,'' \emph{Theoretical Computer Science}, vol. 410, no.~42, pp. 4262--4291, 2009.

\bibitem{dosovitskiy2017carla}
A.~Dosovitskiy, G.~Ros, F.~Codevilla, A.~Lopez, and V.~Koltun, ``{CARLA}: {An} open urban driving simulator,'' in \emph{Conference on Robot Learning (CoRL)}, 2017, pp. 1--16.

\bibitem{li2021metadrive}
Q.~Li, Z.~Peng, Z.~Xue, Q.~Zhang, and B.~Zhou, ``{MetaDrive}: Composing diverse driving scenarios for generalizable reinforcement learning,'' \emph{arXiv preprint arXiv:2109.12674}, 2021.

\bibitem{sloane}
\BIBentryALTinterwordspacing
N.~J.~A. Sloane, ``A001035: Number of partially ordered sets (``posets'') with n labeled elements.'' [Online]. Available: \url{https://oeis.org/A001035}
\BIBentrySTDinterwordspacing

\bibitem{hamming1950error}
R.~W. Hamming, ``Error detecting and error correcting codes,'' \emph{The Bell System Technical Journal}, vol.~29, no.~2, pp. 147--160, 1950.

\bibitem{californiadmv2025report}
\BIBentryALTinterwordspacing
{California Department of Motor Vehicles}, ``Autonomous vehicle collision reports.'' [Online]. Available: \url{https://www.dmv.ca.gov/portal/vehicle-industry-services/autonomous-vehicles/autonomous-vehicle-collision-reports/}
\BIBentrySTDinterwordspacing

\bibitem{elmaaroufi2024scenicnl}
K.~Elmaaroufi, D.~Shanker, A.~Cismaru, M.~Vazquez-Chanlatte, A.~Sangiovanni-Vincentelli, M.~Zaharia, and S.~A. Seshia, ``Scenic{NL}: Generating probabilistic scenario programs from natural language,'' in \emph{1st Conference on Language Modeling (COLM)}, 2024.

\bibitem{comanici2025gemini}
G.~Comanici, E.~Bieber, M.~Schaekermann, I.~Pasupat, N.~Sachdeva, I.~Dhillon, M.~Blistein, O.~Ram, D.~Zhang, E.~Rosen \emph{et~al.}, ``Gemini 2.5: Pushing the frontier with advanced reasoning, multimodality, long context, and next generation agentic capabilities,'' \emph{arXiv preprint arXiv:2507.06261}, 2025.

\bibitem{schulman2017proximal}
J.~Schulman, F.~Wolski, P.~Dhariwal, A.~Radford, and O.~Klimov, ``Proximal policy optimization algorithms,'' \emph{arXiv preprint arXiv:1707.06347}, 2017.

\bibitem{dreossi2019verifai}
T.~Dreossi, D.~J. Fremont, S.~Ghosh, E.~Kim, H.~Ravanbakhsh, M.~Vazquez-Chanlatte, and S.~A. Seshia, ``{VerifAI}: A toolkit for the formal design and analysis of artificial intelligence-based systems,'' in \emph{International Conference on Computer Aided Verification (CAV)}.\hskip 1em plus 0.5em minus 0.4em\relax Springer, 2019, pp. 432--442.

\bibitem{bradley1952rank}
R.~A. Bradley and M.~E. Terry, ``Rank analysis of incomplete block designs: I. the method of paired comparisons,'' \emph{Biometrika}, vol.~39, no. 3/4, pp. 324--345, 1952.

\end{thebibliography}
